\DocumentMetadata{testphase=new-or-1}

\documentclass{article} %
\PassOptionsToPackage{table}{xcolor}
\usepackage{iclr2027_conference,times}

\usepackage{amsmath,amsfonts,bm}

\def\eqref#1{equation~\ref{#1}}

\def\1{\bm{1}}

\DeclareMathAlphabet{\mathsfit}{\encodingdefault}{\sfdefault}{m}{sl}
\SetMathAlphabet{\mathsfit}{bold}{\encodingdefault}{\sfdefault}{bx}{n}

\usepackage[hyperfootnotes=false]{hyperref}
\hypersetup{
    colorlinks=false,       %
    pdfborder={0 0 1},      %
    pdfborderstyle={},     %
    citebordercolor=green  %
}

\usepackage{url}
\usepackage{amsmath,amssymb}
\usepackage{booktabs}
\usepackage{graphicx}
\usepackage{multirow}
\usepackage{tcolorbox}
\usepackage{algorithm}
\usepackage{algpseudocode}
\usepackage{wrapfig}
\floatstyle{plain}\restylefloat{algorithm}

\makeatletter
\newcommand{\minipagecaption}[1]{\def\@captype{table}\caption{#1}}
\makeatother

\newtcolorbox{insbox}{colback=yellow!8, colframe=yellow!55!black,
  boxrule=0.6pt, left=6pt, right=6pt, top=4pt, bottom=4pt, arc=2pt}
\newcommand{\ins}[2]{\begin{insbox}\textbf{Insight #1.} #2\end{insbox}}

\title{Discrete Annotation, Continuous Preference:\\ Rethinking Supervision for Accurate and Generalizable Aesthetic Image Cropping}

\author{
    \noindent Ziqing Zhang$^{1}$\thanks{Equal contribution}~,\enspace
    Xiao Liu$^{1}$\footnotemark[1]~,\enspace
    Kai Liu$^{1}$~,\enspace 
    Jianze Li$^{1}$~,\enspace
    Weihang Zhang$^{2}$~,\enspace\\
    ~\textbf{Linghe Kong}$^{1}$~,\enspace
    \textbf{Yulun Zhang}$^{1}$\thanks{Corresponding authors: Yulun Zhang, yulun100@gmail.com}\\
    \textsuperscript{1}Shanghai Jiao Tong University,\enspace
    \textsuperscript{2}Institute of Media Technology and Experience Design, Huawei
}

\iclrfinalcopy
\definecolor{venuegray}{gray}{0.45}   %
\definecolor{grouprow}{RGB}{238,246,255} %
\definecolor{oursrow}{rgb}{1.0,0.995,0.92}  %
\newcommand{\headshade}{\makebox[0pt][l]{\color{grouprow}\rule[-\dp\strutbox]{\linewidth}{\dimexpr\ht\strutbox+\dp\strutbox\relax}}}
\newcommand{\dashedmidrule}{%
  \noalign{\vskip\aboverulesep}%
  \noalign{\hbox to\linewidth{\leaders\hbox{\rule{2.2pt}{\lightrulewidth}\hskip2.2pt}\hfil}}%
  \noalign{\vskip\belowrulesep}}
\newcommand{\venue}[1]{{\scriptsize\textsf{\textcolor{venuegray}{(#1)}}}}

\begin{document}

\maketitle
\fancyhf{}
\cfoot{\thepage}
\renewcommand{\headrulewidth}{0.4pt}
\renewcommand{\footrulewidth}{0pt}
\vspace{-10mm}
\begin{figure}[!h]
\begin{center}
\includegraphics[width=\textwidth]{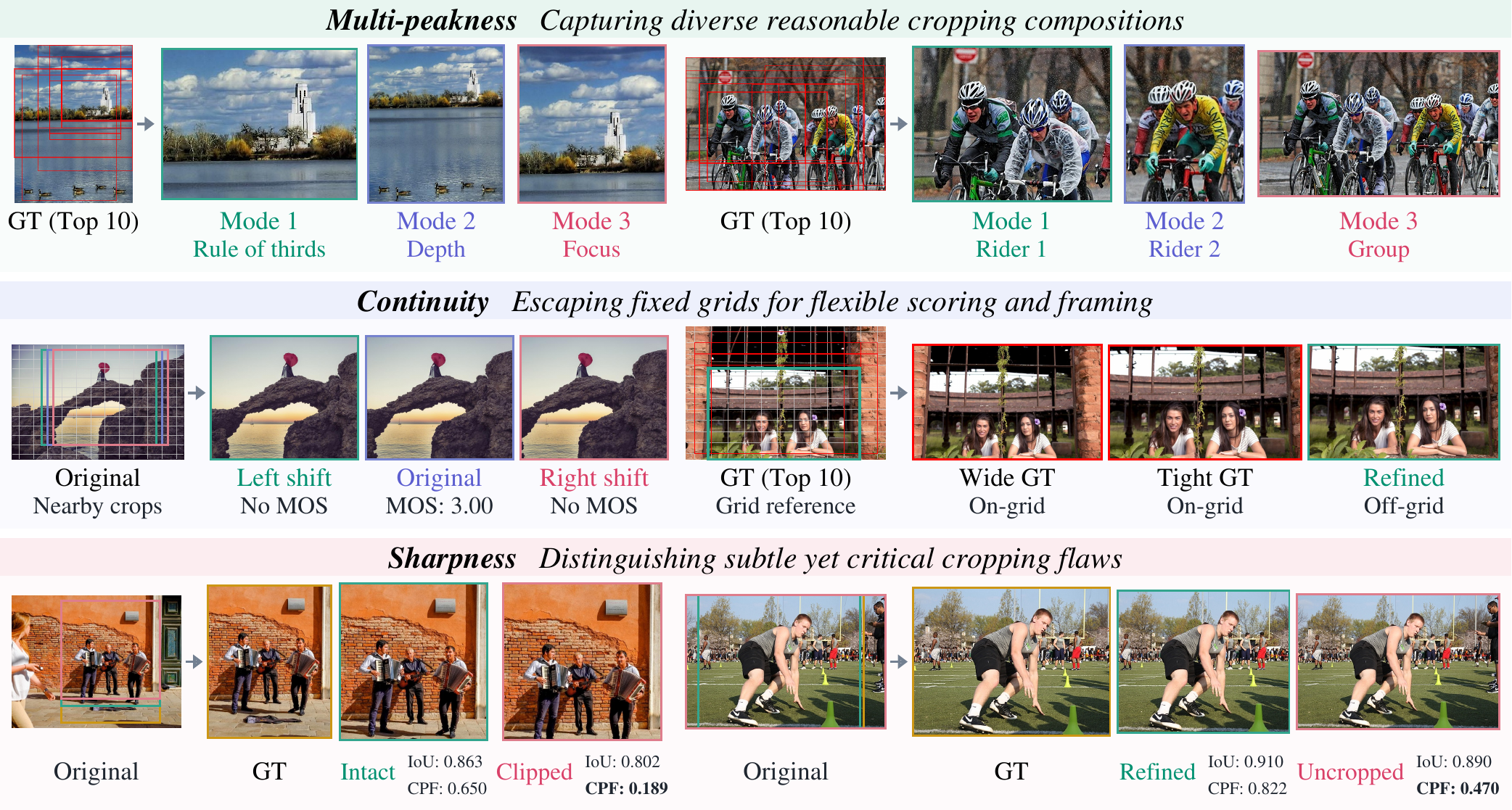}
\end{center}
\vspace{-3mm}
\caption{
\textbf{Illustration of three key properties of our Continuous Preference Field (CPF)}: (1) multi-peakness, (2) continuity, and (3) sharpness. (Zoom-in for best view)}
\label{fig:teaser}
\end{figure}
\vspace{0mm}
\begin{abstract}
\vspace{-2mm}
Aesthetic image cropping aims to identify the optimal crop of an image in terms of aesthetics and composition. While supervision based on annotated data is fundamental, the field has been hindered by a long-standing problem: existing datasets suffer from (1) human subjectivity and (2) rigid discreteness confined to fixed sampling grids. These flawed annotations not only limit the accuracy and generalization of trained models but also severely distort fair evaluation. To overcome this, we propose to model human cropping preference as a \textbf{multi-peaked, continuous, and sharp} field over the crop space. We introduce the \textbf{Continuous Preference Field} (CPF), which recovers a dense preference landscape from discrete annotations through (1) peak clustering, (2) off-lattice refinement, (3) negative shaping, and (4) field assembly. Based on this, we train \textbf{CPIC}, a VLM-based cropping model optimized via GRPO with the CPF reward, which overcomes template collapse, achieving state-of-the-art performance and exceptional out-of-domain generalization. Finally, to resolve the long-standing benchmark evaluation crisis, we introduce \textbf{CPICD}, a comprehensive recalibration of existing ground-truth boxes. By leveraging the CPF to correct grid-bound artifacts across mainstream benchmarks, CPICD establishes a rigorous and reliable foundation for future cropping research. Extensive experiments and user studies demonstrate the superiority of our CPF, CPIC, and CPICD. Code, model, and data are available at \url{https://github.com/zzqingz/CPIC}.
\end{abstract}

\section{Introduction}
\vspace{-2mm}

Aesthetic image cropping is a classic computer vision task that seeks to crop the sub-image from an image with the best aesthetics and composition. It serves critical roles in real-world applications such as automated photo-editing and intelligent camera view recommendation. Over the past decade, solutions have evolved from traditional networks scoring pools of proposals~\citep{yan2013icdb,chen2017vfn,wei2018cpc,zeng2019gaic,zeng2020gaic,zhang2022hcic}, to models directly regressing continuous coordinates~\citep{hong2021cacnet,yang2023sacnet,liu2023unic,zhang2026procrop}, and most recently to vision-language models (VLMs) decoding bounding boxes as discrete text tokens~\citep{lee2025cropper,sheng2025instructcrop,du2026venus}. By leveraging powerful visual pre-training knowledge and advanced reasoning capabilities, VLMs have significantly advanced scene understanding. Nevertheless, despite these architectural leaps, the vast majority of state-of-the-art approaches share a core dependency: to accurately align with human aesthetic preference, they fundamentally rely on fitting the specific data bounding boxes that human annotators scored highest.

\begin{table}[t]
\setlength{\tabcolsep}{3pt}
\begin{minipage}[t]{0.50\textwidth}
\caption{Cropping datasets. Each restricts the coordinate space to make
annotation affordable, and each records the taste of a small pool of people.}
\label{tab:datasets}
\centering\small
\begin{tabular*}{\linewidth}{@{\extracolsep{\fill}}lll@{}}
\toprule
\headshade{}Dataset & Candidates & Subjects \\
\midrule
CPC~\citep{wei2018cpc}    & $4{\times}6$ views & $6$ workers \\
FLMS~\citep{fang2014flms} & $10$ boxes         & $10$ experts \\
FCDB~\citep{chen2017fcdb} & $1$ box            & $7$ workers \\
GAIC~\citep{zeng2019gaic} & $12{\times}12$ grid & $19$ raters \\
\bottomrule
\end{tabular*}
\end{minipage}\hfill
\begin{minipage}[t]{0.47\textwidth}
\caption{Distinct coordinate values and distinct boxes produced over the $200$
GAIC test images: at most $800$ and $200$ respectively.}
\label{tab:gridmethods}
\centering\small\setlength{\tabcolsep}{2pt}
\begin{tabular*}{\linewidth}{@{\extracolsep{\fill}}llrr@{}}
\toprule
\headshade{}Method & Output & Values & Boxes \\
\midrule
HCIC~\citep{zhang2022hcic} & anchors & $18$  & $93$ \\
GAIC~\citep{zeng2019gaic} & anchors & $19$  & $105$ \\
Venus~\citep{du2026venus}  & tokens  & $43$  & $66$ \\
CPIC (ours)                       & tokens  & $\mathbf{157}$ & $\mathbf{199}$ \\
\bottomrule
\end{tabular*}
\end{minipage}
\vspace{-3mm}
\end{table}

However, a critical examination of existing datasets reveals two major flaws. First, they are \textbf{(1) subjective}. As summarized in Table~\ref{tab:datasets}, mainstream datasets record the highly idiosyncratic preferences of a tiny pool of raters (e.g., relying on just 6 to 19 subjects), causing models to exhibit \textbf{poor generalization} across different benchmarks. Second, they are artificially \textbf{(2) discrete}. To make annotation affordable, these datasets strictly confine candidate boxes to sparse grids or predefined aspect ratios. Training on these restricted grids forces models into template collapse, resulting in \textbf{poor spatial accuracy}. As demonstrated in Table~\ref{tab:gridmethods}, Venus~\citep{du2026venus}, for instance, produces only 43 distinct coordinate values across 800 test slots.

How, then, can we break free from these deeply flawed dataset priors? As long as models are trained to strictly mimic the limited, discrete bounding boxes provided by annotators, they will never uncover the true continuous nature of aesthetics. To solve this, we must shift our perspective from discrete fitting to modeling the global aesthetic landscape.

\ins{1}{Aesthetic cropping preference is a multi-peaked, continuous, and sharp distribution.}

As illustrated in Figure~\ref{fig:teaser}, true human preference differs drastically from rigid grid annotations. (1) It is \textit{multi-peaked}: a single image frequently supports multiple disjoint, valid framings. For instance, both a tight portrait crop and a wider rule-of-thirds composition can be equally pleasing. (2) It is \textit{continuous}: shifting a good crop by a few pixels still yields a good crop, demanding sub-grid spatial continuity rather than artificial score drops between grid points. (3) It is \textit{sharp}: however, once a critical visual boundary is crossed (e.g., clipping a subject's face), the aesthetic score plummets steeply—a strict penalty that standard metric smoothing (like IoU) fails to reflect.

If true preference exhibits these complex geometric properties, how can we bridge the gap between this continuous ideal and our discrete dataset reality?

\ins{2}{Discrete annotations can provide reliable anchors to recover continuous landscape.}

Crucially, we do not need to discard decades of existing datasets. While individual grid boxes are discrete and noisy, collectively, they roughly sample the core tendencies of human preference. We can construct a continuous approximation to human preference by using them as starting points.

Guided by these insights, we introduce the \textbf{Continuous Preference Field (CPF)}, constructed via a four-stage pipeline. \textbf{(1) Peak clustering} identifies the multiple valid framings within an image to guarantee the \textit{multi-peaked} property. \textbf{(2) Off-lattice refinement} shifts these targets off the discrete grid to their precise continuous local optima, ensuring \textit{continuity}. \textbf{(3) Negative shaping} injects hard boundary penalties (e.g., truncated subjects) to enforce \textit{sharpness}. \textbf{(4) Field assembly} merges them into a dense, continuous reward function.

With this dense preference landscape established, we introduce \textbf{CPIC}, fundamentally changing the VLM cropping paradigm. Apart from only using standard instruction tuning to regress discrete tokens, CPIC leverages Group Relative Policy Optimization (GRPO)~\citep{shao2024grpo} guided by the CPF reward. This allows CPIC to explore and optimize in the continuous space, completely overcoming template collapse and achieving unmatched out-of-domain generalization.

Finally, the same discrete flaws plaguing training data also compromise testing. Treating current validation ground truths as flawed anchors, we apply our CPF to introduce \textbf{CPICD}. This recalibration of benchmark annotations purges historical grid artifacts, providing the field with its first rigorous continuous evaluation standard.
In summary, our contributions are:
\begin{itemize}\setlength{\itemsep}{0pt}
\item We construct \textbf{Continuous Preference Field} (CPF), which models aesthetic cropping preference as a multi-peaked, continuous, and sharp distribution.
\item We propose \textbf{CPIC}, a VLM-based cropping model trained with CPF as a GRPO reward, which overcomes template collapse to achieve state-of-the-art accuracy.
\item  We introduce \textbf{CPICD}, recalibrating mainstream ground-truth boxes to resolve the long-standing benchmark evaluation crisis.
\item Extensive experiments and user studies  validate the superiority of our framework.
\end{itemize}

\vspace{-4mm}
\section{Related Work}
\vspace{-2mm}

\subsection{Aesthetic image cropping}
\vspace{-2mm}
Aesthetic image cropping is the task of identifying and extracting the most visually appealing~\citep{liu2024dogiqa} and optimally composed sub-region from a given source photograph. Early learned methods approach aesthetic cropping by scoring candidate proposals, either ranking professional crops~\citep{chen2017vfn}, regressing quality changes~\citep{yan2013icdb}, scoring predefined views~\citep{wei2018cpc}, or evaluating dense candidates generated from grid anchors~\citep{zeng2019gaic,zeng2020gaic,zhang2022hcic}. To bypass exhaustive pooling, a second family regresses bounding box coordinates directly, conditioned on various image features~\citep{hong2021cacnet,yang2023sacnet,liu2023unic,zhang2026procrop}. The current state of the art formulates cropping autoregressively, utilizing VLMs to decode bounding boxes as discrete text tokens via prompting, instruction tuning, or pretraining~\citep{lee2025cropper,sheng2025instructcrop,du2026venus}. Because discrete coordinate vocabularies seamlessly encode dataset annotation lattices, these models are highly susceptible to reproducing structural discretization artifacts. In contrast, our method constructs a continuous optimization signal directly from the underlying structure of the original annotations themselves.

\vspace{-2mm}
\subsection{Cropping datasets}
\vspace{-2mm}
Existing cropping datasets often impose strict constraints that artificially limit the valid coordinate space. ICDB initiated the domain with professional crops~\citep{yan2013icdb}, while FCDB retains a single crop preferred over the original image~\citep{chen2017fcdb}, and FLMS collects unordered boxes from crowd workers~\citep{fang2014flms}. To evaluate denser candidates, CPC restricts views to four aspect ratios~\citep{wei2018cpc}, and GAIC strictly confines approximately ninety scored candidates per image to a $12\times12$ anchor grid~\citep{zeng2019gaic,zeng2020gaic}. Other datasets supply subject-centric ranked pairs~\citep{yang2023sacnet} or rely on image-level aesthetic scores for auxiliary supervision without bounding boxes~\citep{murray2012ava}. To avoid subjective human labeling, GenCrop synthesizes training pairs by outpainting professional photographs~\citep{hong2024gencrop}. Rather than discarding human judgments, our approach builds on existing discrete annotations and enables crop evaluation beyond annotated candidates and refinement of existing ground-truth boxes.

\vspace{-2mm}
\section{Methodology}
\vspace{-2mm}

\subsection{Preliminary: Rethinking the Supervision Signal} \label{sec:rethink}
\vspace{-1mm}
Given a source image $I$, aesthetic image cropping predicts an admissible bounding box $b = (x_1, y_1, x_2, y_2)$. Despite its conceptual simplicity, formulating a reliable supervision signal is persistently hindered by two structural flaws in existing benchmarks.

First, aesthetic preference is inherently \textbf{subjective}. As the last column of Table~\ref{tab:datasets} shows, every benchmark rests on a small pool of just 6 to 19 annotators, meaning the mean opinion score (MOS) contains error noise and cannot be absolutely trusted. Additionally, this causes a lack of generalizability from one dataset to another, as the annotators of different datasets may prefer differently.

Second, bounding box evaluation is treated as \textbf{discrete}. The candidate-space column of Table~\ref{tab:datasets} shows that canonical datasets restrict candidate crops to coarse anchor grids or sparse predefined aspect ratios. This discreteness systematically excludes the  unconstrained coordinate space, forcing models to artificially approximate fluid geometric operations. Table~\ref{tab:gridmethods} shows the severe cost of this discreteness. Across two hundred images, published croppers~\citep{zhang2022hcic,zeng2019gaic,du2026venus} emit only 18 to 43 distinct coordinate values, repeating whole boxes across unrelated photographs; the continuous coordinates they cannot produce are simply the ones their supervision never contained, which restricts the performance from reaching its optimal upper bound.

\begin{figure}[t]
\begin{center}
\includegraphics[width=\textwidth]{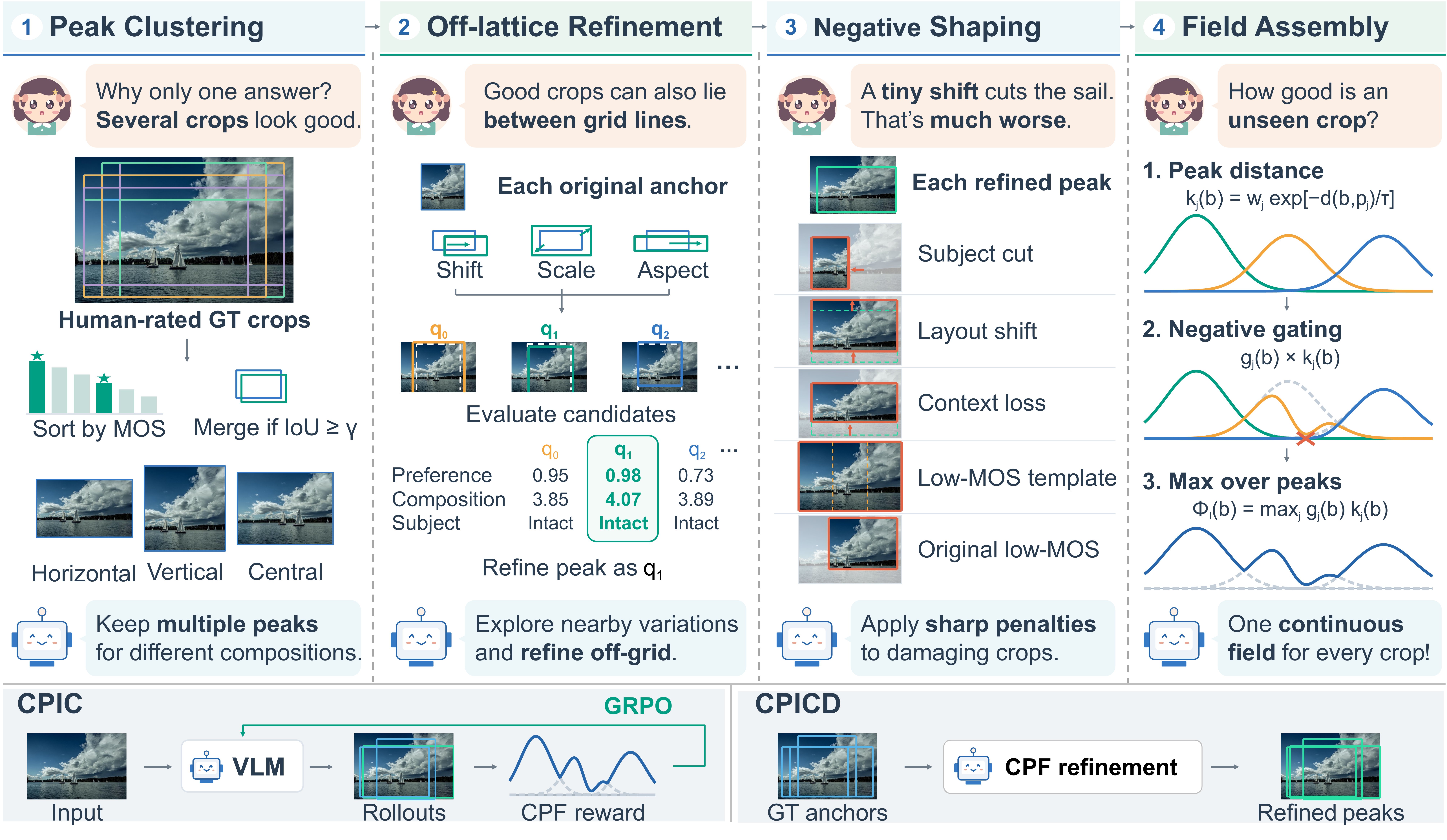}
\end{center}
\vspace{-4mm}
\caption{Overview of CPF construction and its applications. (1) Peak clustering identifies distinct preference modes from human-scored crops. (2) Off-lattice refinement adjusts each peak using preference, composition, and subject-preservation scores. (3) Negative shaping constructs five types of undesirable crops. (4) Field assembly combines peak-distance and negative gating into CPF. CPF serves as the reward for GRPO training of CPIC and guides annotation recalibration for CPICD.}
\vspace{-5mm}
\label{fig:pipeline}
\end{figure}

\vspace{-2mm}
\subsection{CPF: Constructing the Continuous Preference Field} \label{sec:cpf}
\vspace{-2mm}

To bridge the gap between discrete, noisy labels and the human-preferred aesthetic landscape, we introduce the Continuous Preference Field (CPF). Our goal is to explicitly instantiate the three insights identified in the Introduction---multi-peakedness, continuity, and sharpness---into a dense, evaluable scalar function. The four-stage construction is illustrated in Figure~\ref{fig:pipeline}.

\begin{wrapfigure}[11]{r}{0.54\textwidth}
\vspace{-4mm}
\begin{minipage}{0.54\textwidth}
\footnotesize
\refstepcounter{algorithm}\label{alg:peaks}%
\hrule height 0.9pt \vskip 2pt
\noindent\textbf{Algorithm \thealgorithm} Peak Clustering\par
\vskip 2pt \hrule height 0.4pt \vskip 3pt
\begin{algorithmic}[1]
\Require candidates $C$ with scores $s$; $\gamma,\Delta,K,T$
\State sort $C$ by $s$ desc.; $s_{\max}\gets\max s$; $P\gets\emptyset$
\For{$c_j\in C$ \textbf{while} $|P|<K$}
  \State \textbf{if} $s_j < s_{\max}-\Delta$ \textbf{and} $P\neq\emptyset$ \textbf{then continue}
  \State \textbf{if} $\mathrm{IoU}(c_j,p)<\gamma$ for all $p\in P$ \textbf{then} $P\gets P\cup\{c_j\}$
\EndFor
\State \textbf{if} $P=\emptyset$ \textbf{then} $P\gets\{\arg\max_j s_j\}$
\State $w_j\gets\exp\big(-(s_{\max}-s_j)/T\big)$ for each $p_j\in P$
\State \Return peaks $P$ with weights $W$
\end{algorithmic}
\vskip 2pt \hrule height 0.9pt
\end{minipage}
\vspace{-4.5mm}
\end{wrapfigure}
\textbf{Stage 1: Peak Clustering.} To model aesthetic preference, we must capture its inherently \textbf{multi-peaked} distribution. We observe that existing human annotations naturally manifest this property: high-scoring crop candidates concentrate densely around specific compositional modes while remaining sparse elsewhere. This spatial concentration directly embodies the multi-peakedness we seek, which we recover using the clustering procedure in Algorithm~\ref{alg:peaks}. Furthermore, by assigning dynamic weights via the dataset's annotation noise floor $T$, we derive both the number of peaks and their relative weights directly from each specific image's native score spread. Employing a fixed weight ladder would incorrectly impose an identical rank-1 to rank-2 gap across all images, acting as a rigid hyper-parameter that erases inherent aesthetic subjectivity.

\textbf{Stage 2: Off-lattice Refinement.} The peaks extracted in Stage 1 still reside on the quantized annotation grid. To restore \textbf{continuity}, we expand the local search space beyond the discrete lattice. We refine each clustered peak by selecting the best candidate from the original anchor and its continuously perturbed neighbors, combining learned preference with rule-based regularization to reduce subjective annotation bias. 
Around an anchor center $(c_x, c_y)$ with width $w$ and height $h$, deterministic perturbations are seeded via the image key, peak index, and anchor rank:
\begin{equation}
\begin{aligned}
c_x' &= c_x + \mathcal{N}(0, \sigma_{\mathrm{shift}} w), \quad & w' &= w \exp(s + a/2), \\
c_y' &= c_y + \mathcal{N}(0, \sigma_{\mathrm{shift}} h), \quad & h' &= h \exp(s - a/2),
\end{aligned}
\end{equation}
where $s \sim \mathcal{N}(0, \sigma_{\mathrm{scale}})$ and $a \sim \mathcal{N}(0, \sigma_{\mathrm{aspect}})$.
We refine targets using the mean score $z = (z_{\mathrm{pref}} + z_{\mathrm{comp}} + z_{\mathrm{subj}})/3$. The score fuses:
\begin{itemize}
\item \textbf{(1) Human preference ($z_{\mathrm{pref}}$):} Averages the standardized anchor MOS and pre-trained GAIC score~\citep{zeng2019gaic,zeng2020gaic}: $z_{\mathrm{pref}}=(z_{\mathrm{MOS}}+z_{\mathrm{GAIC}})/2$. Because the GAIC scorer learns generalized rules from dataset annotations, it effectively smooths out the sharp noise and discreteness of raw human labels.
\item \textbf{(2) Composition score ($z_{\mathrm{comp}}$):} Provided by SAMP-Net~\citep{zhang2021sampnet}, this robustly evaluates fine-grained sub-grid geometric alignment.
\item \textbf{(3) Subject preservation ($z_{\mathrm{subj}}$):} Computes the explicit area retention percentage of YOLOv8~\citep{jocher2023yolov8}-detected subject bounding boxes.
\end{itemize}
Composition and subject-preservation criteria regularize the learned preference score with explicit visual constraints, reducing reliance on subjective annotations and biases inherited by one scorer. 

\textbf{Stage 3: Negative Shaping.} Spatial distance alone implies a smooth decay, which contradicts the fact that crossing critical visual boundaries severely plummets perceived aesthetics. To enforce this \textbf{sharpness}, we explicitly model "valleys" representing non-preference. We surround each continuous peak with structured negatives encompassing five classes:

\begin{itemize}
\item \textbf{(1) Truncated subject:} We intentionally push the box edges inward to clip the bounding box of YOLOv8~\citep{jocher2023yolov8,hong2024gencrop}-detected main subjects.
\item \textbf{(2) Broken composition:} We shift the box to deliberately move detected subjects away from their original ideal placements (e.g., rule-of-thirds or center) to awkward positions near the frame edges, and shift detected horizon lines away from rule-of-thirds/center axes.
\item \textbf{(3) Discarded value:} We tighten the bounding box edges from all directions without cutting the subject, blindly discarding meaningful contextual background information.
\item \textbf{(4) Templated crops:} We include the uncropped full image and highly repetitive boxes lying on discrete grid points; if these templates do not possess a natively high MOS in the dataset, they are strictly marked as negatives.
\item \textbf{(5) Original low-MOS:} We dynamically mine original human-annotated boxes that highly overlap with the refined peak but possess a demonstrably low MOS.
\end{itemize}
 By explicitly synthesizing these five boundary conditions, we inject necessary sharpness into the continuous space, ensuring the field rigorously models the severe aesthetic penalties humans perceive when fundamental compositional rules are violated.

\begin{figure}[t]
\centering
\input{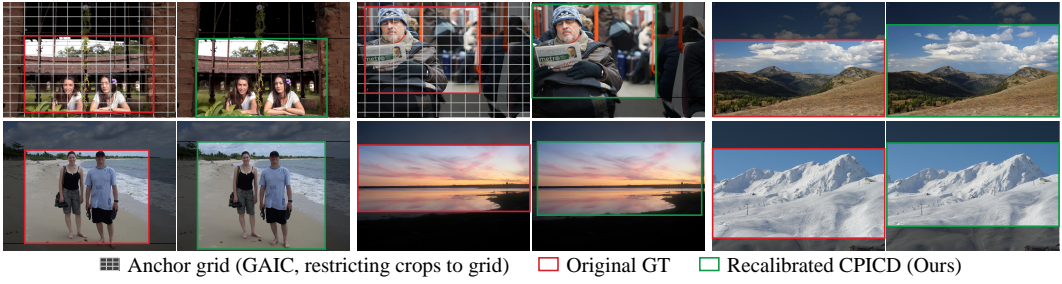}
\vspace{-6mm}
\caption{Comparison of original benchmark annotations and CPICD recalibrated boxes. Recalibration improves unbalanced compositions with truncated subjects or overly narrow crops.}
\vspace{-4mm}
\label{fig:recal}
\end{figure}

\textbf{Stage 4: Field Assembly.} Finally, we unify the peaks, their dynamic weights, and the structured negatives into a continuous reward field. To measure geometric discrepancy without introducing additional weighting hyperparameters, we combine four complementary terms for overlap, boundary displacement, scale, and shape:
\begin{equation}
d(b,b')
=
\frac{1}{4}
\left[
(1-\operatorname{IoU}(b,b'))
+\operatorname{BDE}(b,b')
+\frac{|\log(A/A')|}{1+|\log(A/A')|}
+\frac{|\log(r/r')|}{1+|\log(r/r')|}
\right],
\end{equation}
where $A$ and $r$ denote the area and aspect ratio, respectively, and BDE is computed on normalized coordinates. All four terms are thus bounded to comparable ranges and contribute equally, avoiding manually tuned distance weights.

For each compositional peak $p_j$, its refined location  induces a smoothly decaying similarity score $\exp(-d(b,p_j)/\tau)$, which attracts a query crop $b$ toward the corresponding high-preference region. Spatial proximity alone, however, cannot capture the sharp preference drop near undesirable crops. We therefore compare the distance to the positive peak against that to its nearest structured negative and define a soft negative gate:
\begin{equation}
g_j(b)
=
\sigma\!\left(\bigl[\min_{n\in N_j} d(b,n)-d(b,p_j)\bigr]/\tau\right),
\end{equation}
where $\sigma(\cdot)$ denotes the sigmoid function. The gate remains high when $b$ is closer to the preferred peak, but rapidly decreases as it approaches an undesirable region. The shared temperature $\tau$ controls both the locality of each peak and the sharpness of this transition.

Finally, since an image may support multiple valid framings, a query crop only needs to match one preference mode. We therefore take the maximum over all peak responses, weighted by their relative importance $w_j$:
\begin{equation}
\Phi_I(b)
=
\max_j
\left[
w_j
\exp\!\bigl(-d(b,p_j)/\tau\bigr)
g_j(b)
\right].
\end{equation}
Through this assembly, $\Phi_I(b)$ forms our \textbf{multi-peaked, continuous, and sharp} preference field.

\subsection{CPIC: Leveraging CPF as Supervision Signal} \label{sec:train}
With our continuous and highly distinguishable reward landscape established, we introduce CPIC. We first employ a cold-start supervised fine-tuning stage~\citep{guo2025deepseekr1} using the refined continuous peaks as cross-entropy targets, explicitly shifting the VLM's output support off the original discrete annotation lattice.
Following this cold start, we apply group-relative policy optimization (GRPO) \citep{shao2024grpo}, sampling a group of crops $o_1 \dots o_G$ per image $q$ to optimize:
\begin{equation}
J(\theta) = \mathbb{E}_{q,\,\{o_i\}\sim\pi_{\text{old}}} \left[ \frac{1}{G} \sum_{i=1}^G \left( \min\!\big(\rho_i A_i,\ \text{clip}(\rho_i, 1{-}\epsilon, 1{+}\epsilon) A_i\big) - \beta\,\mathbb{D}_{\text{KL}}(\pi_\theta \| \pi_{\text{ref}}) \right) \right],
\end{equation}
where $\rho_i = \frac{\pi_\theta(o_i | q)}{\pi_{\text{old}}(o_i | q)}$, and the group-relative advantage $A_i = \frac{R_i - \text{mean}(R)}{\text{std}(R)}$ normalizes the reward $R_i = \Phi_I(b_i)$, the CPF score of the box decoded from rollout $o_i$. Specifically, our CPF allows good boxes and bad boxes to have a large degree of distinguishability. Because of this, the reward maintains a healthy gap and does not collapse, thereby driving the model's predicted boxes to further optimize toward better ones.

\subsection{CPICD: Recalibration of Benchmark Ground Truth with CPF} \label{sec:recal}
As established in Section~\ref{sec:rethink}, the quantization errors that hinder model training equally distort benchmark evaluations, penalizing models for predicting plausible off-grid crops instead of matching flawed grids. To resolve this long-standing evaluation crisis, we deploy our CPF refinement machinery (Stage 2) to audit and recalibrate the standard test sets. By treating each historical benchmark annotation as a flawed initial anchor, applying local continuous spatial perturbations, and selecting the refined crop using the same rule-regularized preference score, we quantitatively correct the grid-bound artifacts (yielding the precise green boxes in Figure~\ref{fig:recal}). This recalibration yields \textbf{CPICD}, establishing an alternative set of continuous ground truth for aesthetic cropping evaluation.

\begin{table}[t]
\caption{Comparison on the four benchmarks. Best per column in \textbf{bold}, second \underline{underlined}.}
\label{tab:main}
\begin{center}
\footnotesize
\setlength{\tabcolsep}{1.3pt}
\begin{tabular}{l@{\hspace{4pt}}ccc@{\hspace{4pt}}ccc@{\hspace{4pt}}ccc@{\hspace{4pt}}cc}
\toprule
\multicolumn{1}{c}{\multirow{2}{*}{Method}} & \multicolumn{3}{c}{GAIC} & \multicolumn{3}{c}{FLMS} & \multicolumn{3}{c}{FCDB} & \multicolumn{2}{c}{CPICD} \\
\cmidrule(lr){2-4}\cmidrule(lr){5-7}\cmidrule(lr){8-10}\cmidrule(lr){11-12}
 & ACC$_{10}\uparrow$ & SRCC$\uparrow$ & LAION$\uparrow$ & IoU$\uparrow$ & BDE$\downarrow$ & LAION$\uparrow$ & IoU$\uparrow$ & BDE$\downarrow$ & LAION$\uparrow$ & IoU$\uparrow$ & BDE$\downarrow$ \\
\midrule
\rowcolor{grouprow}\multicolumn{12}{c}{\textit{Traditional networks}} \\
\addlinespace[0.5mm]
VFN \venue{MM'17} & 0.725 & 0.425 & \textbf{5.286} & 0.763 & 0.058 & 5.035 & 0.630 & 0.100 & 4.913 & 0.701 & 0.076 \\
VEN \venue{CVPR'18} & 0.295 & 0.482 & 5.239 & 0.708 & 0.062 & 5.041 & 0.586 & 0.106 & 4.910 & 0.631 & 0.087 \\
VPN \venue{CVPR'18} & 0.405 & 0.495 & 5.271 & 0.825 & 0.040 & \underline{5.078} & 0.658 & 0.088 & \underline{4.933} & 0.726 & 0.067 \\
GAIC \venue{CVPR'19} & \textbf{0.975} & \underline{0.557} & 5.263 & 0.803 & 0.047 & 5.062 & 0.676 & 0.082 & 4.931 & \underline{0.775} & \underline{0.054} \\
CACNet \venue{CVPR'21} & 0.800 & 0.457 & 5.259 & 0.840 & 0.037 & 5.067 & \underline{0.697} & \underline{0.075} & 4.932 & 0.765 & 0.057 \\
HCIC \venue{ECCV'22} & 0.440 & 0.144 & 5.200 & 0.705 & 0.071 & 5.017 & 0.560 & 0.119 & 4.878 & 0.636 & 0.093 \\
UNIC \venue{ICCV'23} & 0.845 & 0.544 & 5.279 & 0.839 & 0.040 & 5.065 & 0.681 & 0.080 & 4.931 & 0.767 & 0.058 \\
\midrule
\rowcolor{grouprow}\multicolumn{12}{c}{\textit{Zero-shot VLMs}} \\
\addlinespace[0.5mm]
Qwen3-VL-2B & 0.290 & 0.285 & 5.137 & 0.804 & 0.049 & 5.020 & 0.566 & 0.119 & 4.863 & 0.649 & 0.094 \\
Qwen3-VL-4B & 0.215 & 0.261 & 5.029 & 0.696 & 0.063 & 4.905 & 0.507 & 0.130 & 4.779 & 0.550 & 0.113 \\
Qwen3-VL-8B & 0.250 & 0.295 & 5.141 & 0.736 & 0.056 & 5.019 & 0.572 & 0.107 & 4.863 & 0.628 & 0.086 \\
Qwen2.5-VL-7B & 0.185 & 0.097 & 4.920 & 0.656 & 0.077 & 4.871 & 0.390 & 0.171 & 4.638 & 0.494 & 0.131 \\
LLaVA-1.5-7B & 0.065 & 0.114 & 4.603 & 0.428 & 0.134 & 4.570 & 0.365 & 0.178 & 4.504 & 0.356 & 0.169 \\
InternVL3.5-2B & 0.385 & 0.321 & 5.282 & 0.817 & 0.047 & 5.041 & 0.634 & 0.102 & 4.930 & 0.704 & 0.081 \\
\midrule
\rowcolor{grouprow}\multicolumn{12}{c}{\textit{VLM-based cropping methods}} \\
\addlinespace[0.5mm]
InstructCrop \venue{MM'25} & 0.630 & 0.326 & 5.238 & 0.761 & 0.055 & 5.040 & 0.652 & 0.087 & 4.910 & 0.709 & 0.069 \\
Venus \venue{CVPR'26} & \underline{0.945} & 0.541 & 5.270 & \textbf{0.856} & \textbf{0.032} & 5.076 & 0.683 & 0.079 & 4.933 & 0.763 & 0.057 \\
\addlinespace[0.5mm]
\rowcolor{oursrow}\textbf{CPIC (ours)} & 0.905 & \textbf{0.590} & \underline{5.283} & \underline{0.842} & \underline{0.037} & \textbf{5.082} & \textbf{0.714} & \textbf{0.070} & \textbf{4.934} & \textbf{0.789} & \textbf{0.049} \\
\bottomrule
\end{tabular}
\end{center}

\vspace{-3pt}
\setlength{\tabcolsep}{4pt}
\begin{minipage}[t]{0.63\textwidth}
\caption{User study on cropping quality (\%): the fraction of trials in which
each method's crop is chosen as the best one.\strut}
\label{tab:us_crop}
\centering\small
\begin{tabular*}{\linewidth}{@{\extracolsep{\fill}}cccccc@{}}
\toprule
\headshade{}GAIC & CACNet & UNIC & InstructCrop & Venus & \textbf{CPIC (ours)} \\
\midrule
32.7 & 37.2 & \underline{42.0} & 22.8 & 32.5 & \textbf{46.3} \\
\bottomrule
\end{tabular*}
\end{minipage}\hfill
\begin{minipage}[t]{0.34\textwidth}
\caption{User study on ground truth recalibration (\%).\strut}
\label{tab:us_recal}
\centering\small
\begin{tabular*}{\linewidth}{@{\extracolsep{\fill}}cc@{}}
\toprule
\headshade{}Original GT & \textbf{CPICD (ours)} \\
\midrule
35.5 & \textbf{64.5} \\
\bottomrule
\end{tabular*}
\end{minipage}
\vspace{-4mm}
\end{table}

\section{Experiments}
\vspace{-2mm}

\subsection{Experimental Settings}
\vspace{-1mm}

\textbf{Datasets.} We train our model on the CPC~\citep{wei2018cpc} and GAIC~\citep{zeng2019gaic,zeng2020gaic} datasets, optimized via our CPF. Following prior work~\citep{zeng2020gaic,zhang2026procrop}, we evaluate on the GAIC test split, FLMS~\citep{fang2014flms}, and FCDB~\citep{chen2017fcdb}.

\textbf{Metrics.} We comprehensively consider multiple metrics. Following standard work~\citep{zeng2020gaic,zhang2026procrop}, we report the Spearman rank correlation coefficient (SRCC) and $\text{ACC}_{10}$ on the GAIC dataset. For FLMS and FCDB, we report the Intersection over Union (IoU) and the boundary displacement error (BDE)~\citep{chen2017vfn}. We further introduce the no-reference LAION-Aesthetics score~\citep{schuhmann2022laion} to measure the aesthetic quality of the crop box.

\textbf{Implementation details.} We train our CPIC model based on Qwen3-VL-2B-Instruct~\citep{bai2025qwen3vl} using LoRA~\citep{hu2022lora} with a rank of 64 and the AdamW optimizer~\citep{loshchilov2019adamw}. In the SFT stage, we train with a learning rate of $1 \times 10^{-4}$ and batch size of 16 for 2500 steps on a single RTX 4090 GPU. In the GRPO~\citep{shao2024grpo} stage, we train with a learning rate of $3 \times 10^{-5}$ and batch size of 16 for 1200 steps on 4 RTX 4090 GPUs. We sample $G=8$ crops per image and set $\beta=0.04$ and  $\epsilon=0.2$.

\vspace{-2mm}
\subsection{Comparison with the state-of-the-art methods}
\vspace{-2mm}

\textbf{Compared methods.} We compare against three families. \emph{Traditional networks}
that score or regress boxes without a large VLM: VPN~\citep{wei2018cpc},
GAIC~\citep{zeng2019gaic,zeng2020gaic}, CACNet~\citep{hong2021cacnet},
HCIC~\citep{zhang2022hcic} and UNIC~\citep{liu2023unic}. \emph{Zero-shot VLMs} prompted to emit a crop box without any cropping-specific training: Qwen3-VL at $2$B, $4$B and $8$B~\citep{bai2025qwen3vl}, Qwen2.5-VL-7B~\citep{bai2025qwen25vl}, LLaVA-1.5-7B~\citep{liu2024llava} and InternVL3.5-2B~\citep{wang2025internvl}. \emph{VLM-based cropping methods} that fine-tune a vision-language model for the task: InstructCrop~\citep{sheng2025instructcrop}, Venus~\citep{du2026venus} and our CPIC.

\textbf{Quantitative comparison.} As detailed in Table~\ref{tab:main}, CPIC demonstrates the strongest generalization across all three method families and under both reference-based and no-reference metrics, ranking first or second on ten of the eleven columns. On the GAIC benchmark, GAIC  and Venus achieve the top ACC$_{10}$ scores due to their grid priors, as this metric inherently rewards reproducing a lattice-bound annotation. Nevertheless, CPIC yields higher absolute crop quality on the same dataset, outperforming both grid-bound methods on the SRCC and LAION metrics. Crucially, CPIC maintains dominant performance on the FLMS and FCDB benchmarks. Because neither FLMS nor FCDB appears in our training data, these test sets are entirely out of domain for us. Succeeding in these unseen environments provides the strongest evidence of our method's robust generalization.

\begin{figure}[t]
\centering
\input{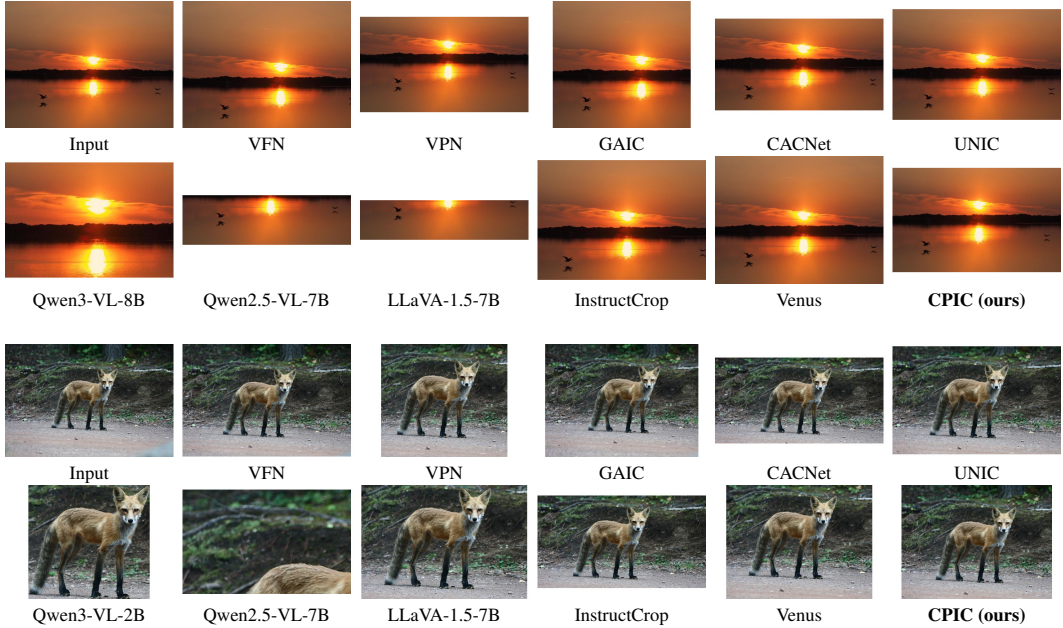}
\vspace{-4mm}
\caption{Qualitative results on in-domain (GAIC) and out-of-domain (FCDB) images.}
\vspace{-6mm}
\label{fig:qualitative}
\end{figure}

\textbf{Qualitative comparison.} Figure~\ref{fig:qualitative} demonstrates our method's aesthetic cropping capabilities compared to baselines. In the first example, CPIC achieves a horizontal, centered composition that perfectly balances the two flying birds on the left and right. Other methods fail to capture this harmony: GAIC, InstructCrop, and Venus leave excessive empty sky and water; VPN and CACNet yield overly thin wide strips. In the second example, CPIC keeps the main subject (the fox) centered while preserving an appropriate margin of context to maintain a harmonious visual balance. Conversely, baseline models either retain distracting empty ground (UNIC, VFN, InstructCrop), crop too tightly around the fox without sufficient breathing room.

\textbf{User Study.} We conduct two users studies with
 implementation details provided in Appendix Sec.~\ref{sec:supp_user}. In the first study evaluating cropping quality (Table~\ref{tab:us_crop}), participants are presented with crops from six methods and instructed to select their top 1 to 3 aesthetically pleasing choices per image. Our CPIC achieves the highest overall selection rate, demonstrating its superior alignment with human visual preference. In the second study evaluating ground-truth recalibration (Table~\ref{tab:us_recal}), participants perform a two-alternative choice between the original benchmark annotation and our recalibrated CPICD. The results show a significant preference for CPICD, verifying that our continuous refinement can correct historical annotation flaws and yield better aesthetic optima. 

\vspace{-1.5mm}
\subsection{Ablation Studies}
\vspace{-1mm}

\begin{table}[t]
\caption{Ablation studies.
Boxes counts the distinct crops the model emits over the $200$ GAIC test images
(at most $200$), the diversity measure of Table~\ref{tab:gridmethods}.}
\label{tab:ablation}
\begin{center}
\small
\setlength{\tabcolsep}{3.2pt}
\begin{tabular*}{\linewidth}{@{\extracolsep{\fill}}llcccccccc@{}}
\toprule
 & \multirow{2}{*}{Configuration} & \multicolumn{3}{c}{CPF ingredient} & GAIC & \multicolumn{2}{c}{FLMS} & FCDB & \multirow{2}{*}{Boxes$\uparrow$} \\
\cmidrule(lr){3-5}\cmidrule(lr){6-6}\cmidrule(lr){7-8}\cmidrule(lr){9-9}
 & & multi-peak & continuity & sharpness & SRCC$\uparrow$ & IoU$\uparrow$ & LAION$\uparrow$ & IoU$\uparrow$ & \\
\midrule
\rowcolor{grouprow}\multicolumn{10}{c}{\textit{Supervision strategies}} \\
\addlinespace[0.5mm]
(a) & CE on annotated GT             &              &              &              & 0.523 & 0.798 & 5.060 & 0.642 & 56 \\
(b) & \quad + GRPO w/ IoU            &              &              &              & 0.564 & 0.822 & 5.074 & 0.671 & 92 \\
(c) & CE on refined peaks            &              & $\checkmark$ &              & 0.538 & 0.827 & 5.076 & 0.704 & 154 \\
(d) & \quad + GRPO w/ IoU            &              & $\checkmark$ &              & 0.559 & 0.835 & \textbf{5.082} & 0.701 & \underline{199} \\
\dashedmidrule
\rowcolor{grouprow}\multicolumn{10}{c}{\textit{+ GRPO with CPF}} \\
\addlinespace[0.5mm]
(e) & CPF w/ single peak             &              & $\checkmark$ & $\checkmark$ & \textbf{0.619} & 0.829 & \underline{5.079} & 0.706 & \underline{199} \\
(f) & CPF w/o refinement             & $\checkmark$ &              & $\checkmark$ & 0.517 & 0.829 & 5.065 & 0.672 & 88 \\
(g) & CPF w/o negatives              & $\checkmark$ & $\checkmark$ &              & 0.574 & \underline{0.839} & 5.076 & \underline{0.708} & \textbf{200} \\
\addlinespace[0.5mm]
\rowcolor{oursrow}(h) & \textbf{CPF (ours)}            & $\checkmark$ & $\checkmark$ & $\checkmark$ & \underline{0.590} & \textbf{0.842} & \textbf{5.082} & \textbf{0.714} & \underline{199} \\
\bottomrule
\end{tabular*}
\end{center}
\vspace{-5mm}
\end{table}

 To systematically validate our design choices, we conduct extensive ablation studies in Table~\ref{tab:ablation}.

\textbf{Effect of the continuous preference field.} We first establish that substituting cross-entropy (CE) with GRPO is not inherently responsible for the success. As shown in \textbf{(a)} and \textbf{(b)}, applying GRPO with a naive IoU reward against discrete ground truths severely limits spatial exploration, capping output diversity at only 92 distinct boxes. While shifting CE targets to our refined continuous peaks \textbf{(c)} expands diversity, the accuracy remains limited. Even when applying GRPO with an IoU reward on these refined continuous targets \textbf{(d)}, the model fails to capture true aesthetic boundaries, yielding sub-optimal results. The substantial, comprehensive improvements across all reference-based and no-reference metrics materialize only when the policy is guided by our CPF \textbf{(h)}.

\textbf{Effect of multi-peak supervision.} Row \textbf{(e)} restricts the CPF to a single peak. Because a single photograph frequently supports several disjoint valid framings, collapsing the field to one exclusive target discards valid alternative crops and forces the model to mimic a single subjective choice. Consequently, while \textbf{(e)} slightly overfits to the GAIC training domain, it artificially penalizes good alternative predictions, suffering noticeable drops in out-of-domain generalization on the FLMS and FCDB benchmarks. By incorporating multi-peak supervision, our full model \textbf{(h)} better aligns with diverse human consensus, yielding a robust balance across all domains.

\textbf{Effect of off-lattice continuity.} Row \textbf{(f)} completely ablates the off-lattice refinement, forcing the CPF positive anchor to revert to the old lattice-bound annotated box. This explicitly destroys the continuity of the target space. As a result, the model's spatial precision collapses, and its ability to break the discrete token constraint is crippled, causing output diversity to plummet back to a mere 88 distinct boxes. Taking the target mathematically off the annotation grid in \textbf{(h)} provides an accurate continuous optimum, allowing the model to freely explore sub-grid coordinates (reaching 199 distinct boxes) and significantly boosting overlap metrics.

\textbf{Effect of sharp structured negatives.} Row \textbf{(g)} ablates the five structured negative classes. Without these negatives, the preference reward decays smoothly and indiscriminately with geometric distance, diluting the learning signal. It fails to aggressively penalize critical aesthetic violations, such as slightly cutting into a subject's face, because the IoU shift is minor. Explicitly gating the field with structured negatives in \textbf{(h)} ensures that preference falls off sharply into a "valley" when visual boundaries are crossed. This sharpness penalty is crucial for fine-grained aesthetic discrimination, maximizing both the no-reference aesthetic quality (LAION) and bounding box accuracy.

Due to page limits, key validity justifications are deferred to Appendix Sec.~\ref{sec:supp_justification}, which we kindly hope will be taken into account for a proper interpretation of our method and results.

\vspace{-2mm}
\section{Conclusion}
\vspace{-3mm}
In this work, we rethink and overcome the long-standing limitations of discrete, grid-bound annotations in aesthetic image cropping. We reframe human preference as a multi-peaked, continuous, and sharp distribution, introducing the Continuous Preference Field (CPF) to transform coarse dataset labels into a dense reward landscape through peak clustering, off-lattice refinement, structured negative shaping, and field assembly. By utilizing the CPF as a GRPO reward, we train CPIC, a VLM-based cropping model optimized directly for this continuous space. CPIC decisively escapes the discrete annotation lattice and achieves state-of-the-art accuracy alongside exceptional out-of-domain generalization. Furthermore, we apply the CPF machinery to audit existing mainstream validation sets, introducing \textbf{CPICD} to recalibrate flawed ground-truth boxes. By correcting the historical grid-bound artifacts that have long distorted fair model comparison, CPICD establishes a rigorous, continuous, and reliable foundation for future aesthetic cropping research.

\newpage

\subsection*{AI use statement}

In this work, we used generative AI tools for assist with translation. We have not used generative AI tools for things like generating data and supporting qualitative and thematic data analysis, and the rest of the required disclosure tasks like mathematical proofs are not applicable to this work. Additionally, we used generative AI tools for polishing our language. We have reviewed all AI-assisted work. We checked LLM-generated research ideas for potential plagiarism through a manual literature survey. LLM-generated code was verified and tested for correctness by all of the authors. We take responsibility for the final content of this work, including text, claims or artifacts produced with the aid of generative AI.

\bibliography{iclr2027_conference}
\bibliographystyle{iclr2027_conference}

\newpage

\appendix
\section*{Appendix}

This supplementary material provides additional implementation details, extended ablations, and discussions to support the main manuscript. It is organized as follows:
\begin{itemize}
    \item Sec.~\ref{sec:supp_inference} summarizes implementation hyperparameters and inference details.
    \item Sec.~\ref{sec:supp_zeroshot} demonstrates CPIC's zero-shot interpretable and interactive cropping capabilities.
    \item Sec.~\ref{sec:supp_justification} provides additional justifications and ablation studies.
    \item Sec.~\ref{sec:supp_user} details the protocols and participant demographics for the human evaluations.
    \item Sec.~\ref{sec:supp_vis} presents additional qualitative visualizations.
\end{itemize}

\section{Implementation details}
\label{sec:supp_inference}
\textbf{Hyperparameters.} We summarize the construction and training settings below.
\begin{itemize}
    \setlength{\itemsep}{2pt}
    \item \textbf{Peak clustering.} We retain at most $K=4$ peaks, with a candidate MOS tolerance $\Delta=0.75$ and initial clustering threshold $\gamma=0.72$. Final peak consolidation uses an IoU threshold of $0.70$. Peak weights use $T=0.35$ in Algorithm~\ref{alg:peaks}.
    \item \textbf{Local perturbations.} We sample $40$ local perturbations per peak. The three $(\sigma_{\mathrm{shift}},\sigma_{\mathrm{scale}},\sigma_{\mathrm{aspect}})$ settings are $(0.012,0.022,0.025)$, $(0.026,0.045,0.050)$, and $(0.050,0.075,0.075)$, sampled with probabilities $0.4$, $0.4$, and $0.2$. 
    \item \textbf{Scoring and negatives.} The three visual scores are averaged equally, with MOS and GAIC scores equally combined within the preference term. The drift penalty uses $\lambda=0.15$, and the field uses a shared temperature $\tau=0.10$. Negatives with IoU $\geq0.85$ to their associated refined peak are removed to avoid conflicting supervision.
    \item \textbf{Optimization.} LoRA uses rank $64$ and alpha $32$, with a warmup ratio of $0.03$. SFT and GRPO use learning rates $10^{-4}$ and $3\times10^{-5}$, respectively, and an effective batch size of $16$. GRPO uses $G=8$, $\beta=0.04$, and $\epsilon=0.2$.
    \item \textbf{Training rollouts.} Crops are sampled with temperature $1.0$, top-$p=0.95$, and a maximum of $64$ generated tokens per rollout.
\end{itemize}

\textbf{Inference details.} At inference time, CPIC predicts a single aesthetic crop for each image using a fixed instruction:

\begin{insbox}
\small
\texttt{Please suggest the single best aesthetic crop region for this image. Use normalized coordinates in a 0-1000 image coordinate system: left edge x=0, right edge x=1000, top edge y=0, bottom edge y=1000. Return only JSON in this exact format: \{"bbox\_2d": [x1, y1, x2, y2]\}.}
\end{insbox}

The prompt defines a normalized coordinate system in which the left and right image boundaries are x=0 and x=1000, while the top and bottom boundaries are y=0 and y=1000. Each crop is represented by four coordinates in the JSON field \texttt{"bbox\_2d"}. We use greedy decoding without sampling, so inference is deterministic and does not require a random seed. All evaluations use single-precision floating-point arithmetic to ensure bit-reproducible results across hardware architectures.

\begin{figure}[!t]
\centering
\colorbox{red!9}{\parbox{0.97\linewidth}{\centering\textbf{CPIC's Interpretable Aesthetic Cropping}}}
\vspace{4pt}

\begin{minipage}[t]{0.215\linewidth}
  \centering
  \textbf{Original photo}\par\vspace{2pt}
  \includegraphics[width=\linewidth,height=3.0cm,keepaspectratio]{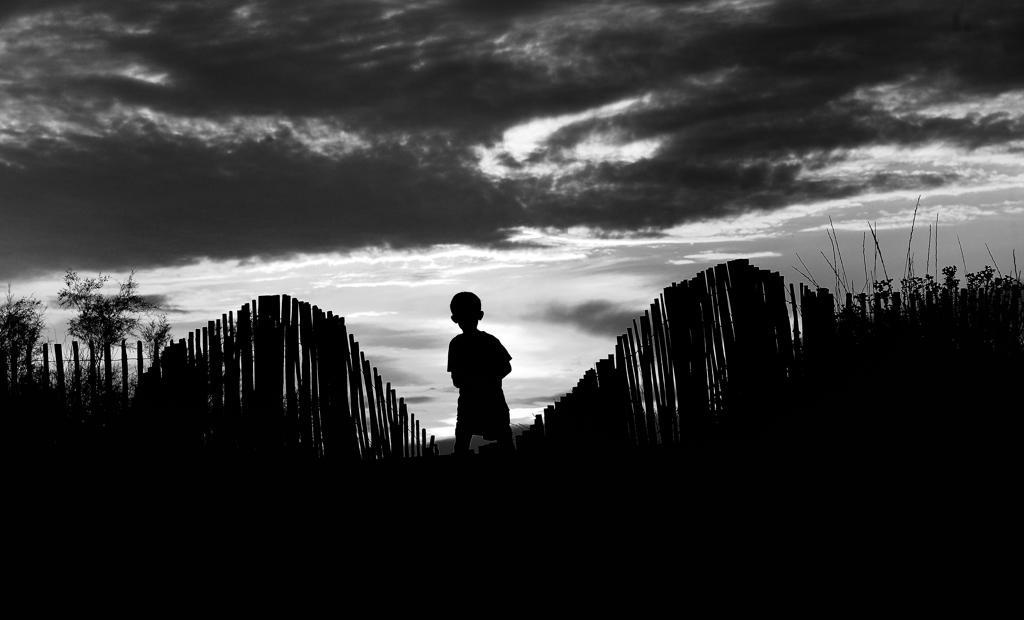}
\end{minipage}
\hfill
\begin{minipage}[t]{0.215\linewidth}
  \centering
  \textbf{CPIC crop}\par\vspace{2pt}
  \includegraphics[width=\linewidth,height=3.0cm,keepaspectratio]{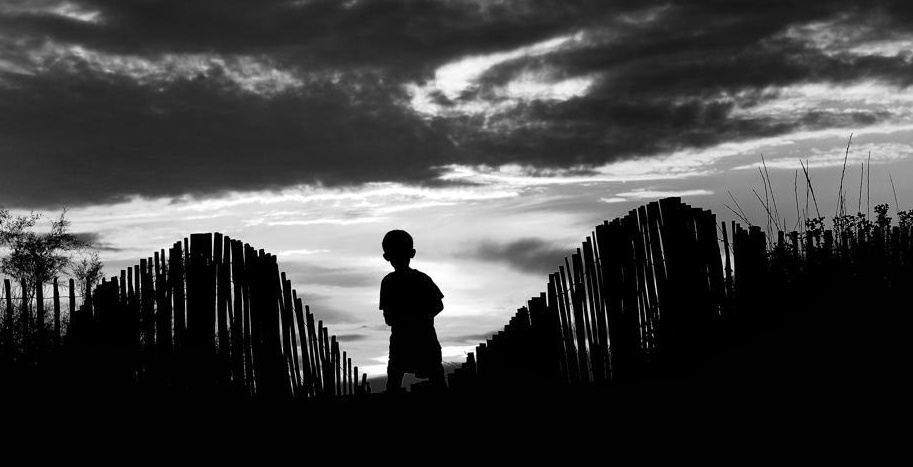}
\end{minipage}
\hfill
\begin{minipage}[t]{0.53\linewidth}
  \small
  \colorbox{blue!6}{\parbox{0.95\linewidth}{\textbf{User:} Give a specific aesthetic rationale for this crop, referring to visible elements and their spatial arrangement. Avoid generic claims. Use two sentences only.}}
  \vspace{4pt}

  \colorbox{blue!7}{\parbox{0.95\linewidth}{\textbf{CPIC:} The crop \textbf{centers} on the silhouette of a child standing \textbf{between two wooden fences}, creating a strong sense of isolation and contemplation. The \textbf{dramatic contrast} between the dark foreground and the bright, textured sky enhances the emotional depth and visual tension of the scene.}}
\end{minipage}

\vspace{9pt}
\colorbox{blue!8}{\parbox{0.97\linewidth}{\centering\textbf{CPIC's Interactive Aesthetic Cropping}}}
\vspace{4pt}

\begin{minipage}[t]{0.215\linewidth}
  \vspace{0pt}\centering
  \small\textbf{Original photo}\par\vspace{2pt}
  \includegraphics[width=\linewidth,height=3.0cm,keepaspectratio]{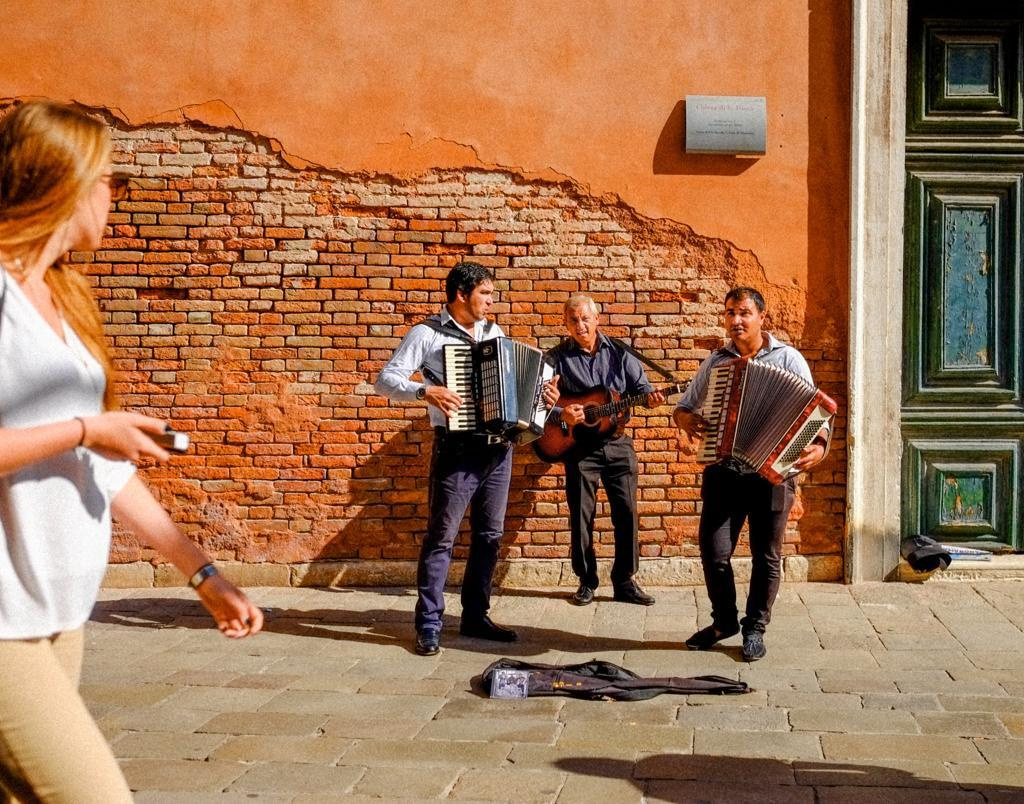}
\end{minipage}
\hfill
\begin{minipage}[t]{0.215\linewidth}
  \vspace{0pt}\centering
  \small\textbf{Turn 1: CPIC crop}\par\vspace{2pt}
  \includegraphics[width=\linewidth,height=3.0cm,keepaspectratio]{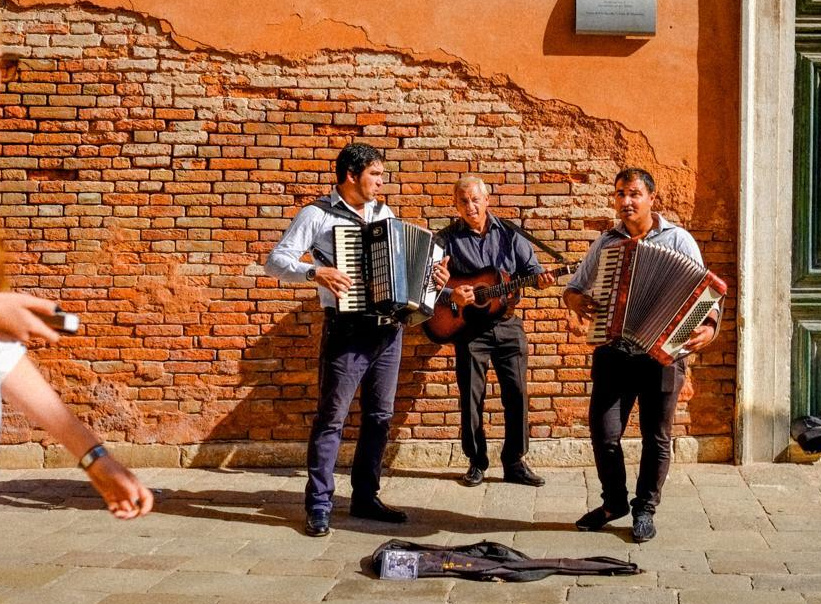}
\end{minipage}
\hfill
\begin{minipage}[t]{0.285\linewidth}
  \vspace{0pt}\centering
  \small\mbox{\textbf{Natural-language feedback}}\par\vspace{5pt}
  \footnotesize The woman at the far left is distracting. Exclude her completely and tightly frame the three musicians.
\end{minipage}
\hfill
\begin{minipage}[t]{0.215\linewidth}
  \vspace{0pt}\centering
  \small\textbf{Turn 2: refined crop}\par\vspace{2pt}
  \includegraphics[width=\linewidth,height=3.0cm,keepaspectratio]{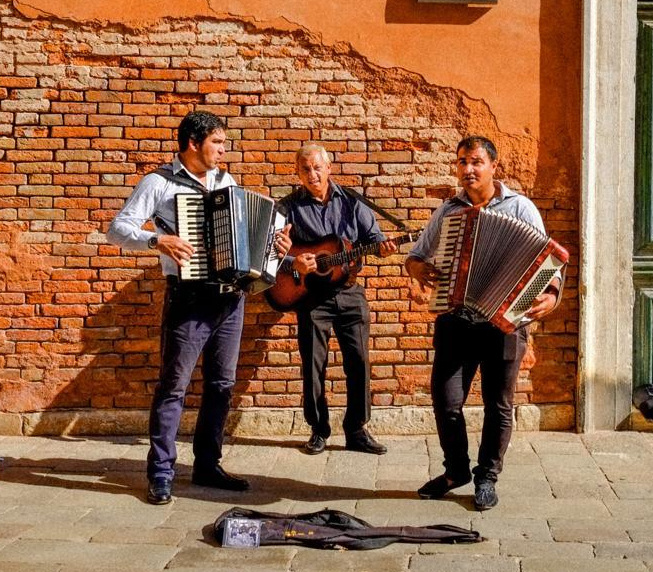}
\end{minipage}

\caption{\textbf{Zero-shot interpretable and interactive aesthetic cropping.} Top: CPIC provides a concrete rationale grounded in the selected subject, spatial arrangement, contrast, and visual effect. Bottom: after its initial crop, CPIC follows free-form feedback to remove the distracting foreground pedestrian and tightly reframe the three musicians. }
\label{fig:supp_text_interaction}
\end{figure}

\section{Zero-shot interpretability and interaction}
\label{sec:supp_zeroshot}
\begin{wraptable}{r}{0.48\textwidth}
\vspace{-\dimexpr\intextsep+\abovecaptionskip+0.52pt\relax}
\vspace{-2mm}
\small
\caption{Output-target ablation. Best \textbf{bolded}.}
\vspace{1.2mm}
\label{tab:supp_short_text}
\setlength{\tabcolsep}{3pt}%
\renewcommand{\arraystretch}{0.88}%
\begin{tabular*}{\linewidth}{@{\extracolsep{\fill}}lccc@{}}
\toprule
\multirow{2}{*}{Output target} & GAIC & FLMS & FCDB \\
\cmidrule(lr){2-2}\cmidrule(lr){3-3}\cmidrule(lr){4-4}
 & ACC$_{10}\uparrow$ & IoU$\uparrow$ & IoU$\uparrow$ \\
\midrule
Box only & 0.905 & \textbf{0.842} & 0.714 \\
Short text + box & \textbf{0.910} & 0.822 & \textbf{0.723} \\
\bottomrule
\end{tabular*}
\vspace{-0.75\baselineskip}
\vspace{-1mm}
\end{wraptable}
We compared box-only output with a variant that first emits a short, image-specific statement about the subject and composition while retaining crop localization as the task objective. Table~\ref{tab:supp_short_text} shows no consistent benefit: the added text improves GAIC and FCDB but degrades FLMS, with only small changes overall. We therefore train CPIC to output only boxes, isolating CPF supervision and avoiding unnecessary generation overhead. This agrees with COMEX~\citep{yang2026comex}, where removing the semantic explanation reward changes crop IoU only marginally. Longer, explicitly grounded chains of thought~\citep{shi2026drscaffold} remain an important direction for future work.\par

Beyond predicting accurate boxes, a practical VLM-based cropper should be able to explain its decisions and respond to a user's intent. Following the qualitative protocol of Venus~\citep{du2026venus}, we therefore probe CPIC in two settings: (1) after its initial crop, we ask why the crop improves the overall aesthetics; and (2) we provide a natural-language follow-up requesting a change in subject, framing, or crop tightness. Importantly, CPIC is not trained with aesthetic-rationale annotations or multi-turn crop-refinement dialogues. These abilities are evaluated directly, without demonstrations or parameter updates, using the same checkpoint as our quantitative evaluation.

Figure~\ref{fig:supp_text_interaction} shows both capabilities with one representative example each. For interpretation, CPIC grounds its explanation in the selected subject, spatial arrangement, tonal contrast, and their visual effect. For interaction, its predicted coordinates respond coherently to free-form feedback by removing an unwanted foreground subject and reframing the requested musicians. Thus, CPF-based crop optimization preserves the underlying model's language capabilities while coupling them to continuous spatial decisions, enabling interpretable and personalized refinement in a zero-shot manner.

\begingroup
\raggedbottom
\section{Additional justification}
\label{sec:supp_justification}

Due to the main-paper page limit, we defer several important discussions and additional justifications to this section. It contains validity analyses that are necessary for a proper assessment of several central claims in this paper, rather than merely additional implementation details. We therefore ask that these results be considered jointly with the main-paper experiments when evaluating the soundness of the proposed framework.

\begin{wraptable}{r}{0.48\textwidth}
\vspace{-4.5mm}
\vspace{-\dimexpr\intextsep+\abovecaptionskip+0.5pt\relax}
\small
\caption{Oracle IoU on three benchmarks.}
\vspace{1.2mm}
\label{tab:supp_coordinate_oracle}
\setlength{\tabcolsep}{2pt}%
\renewcommand{\arraystretch}{0.95}%
\begin{tabular*}{\linewidth}{@{\extracolsep{\fill}}lccc@{}}
\toprule
Coordinates & GAIC & FLMS & FCDB \\
\midrule
$12\times12$ anchors & \textbf{1.0000} & 0.9099 & 0.8273 \\
$0$--$1000$ integers & 0.9994 & \textbf{0.9997} & \textbf{0.9989} \\
\bottomrule
\end{tabular*}
\vspace{-0.75\baselineskip}
\vspace{-1mm}
\end{wraptable}
\textbf{Justification for continuity.} Continuous crop optimization requires a coordinate space dense enough to express precise boundary adjustments. A fixed anchor set limits the attainable accuracy even when the best candidate is selected perfectly. Table~\ref{tab:supp_coordinate_oracle} quantifies this limit by selecting the highest-IoU representable box for each test image under the benchmark's ground-truth matching protocol. GAIC's $12\times12$ anchor template reaches an oracle IoU of 1 on GAIC because its annotations use the same template, but only 0.9099 and 0.8273 on FLMS and FCDB. To remain consistent with the model's pretrained coordinate format, we represent crop boundaries as integers in $[0,1000]$. Although this representation is not strictly continuous, \textbf{its quantization error is negligible}: the near-unit oracle IoU across all three datasets shows that it substantially raises the attainable accuracy beyond the original grid anchors on FLMS and FCDB. Therefore, it is CPF that provides supervision beyond the annotation grid that enables the model to move away from fixed crop templates; we do not claim that the model itself produces strictly continuous coordinates.\par

\begin{wraptable}{r}{0.48\textwidth}
\vspace{-\dimexpr\intextsep+\abovecaptionskip+0.52pt\relax}
\small
\caption{Refinement-score ablation.}
\vspace{1.2mm}
\label{tab:supp_refinement_scores}
\footnotesize
\setlength{\tabcolsep}{1pt}%
\renewcommand{\arraystretch}{0.95}%
\begin{tabular*}{\linewidth}{@{\extracolsep{\fill}}llcccccc@{}}
\toprule
\multirow{2}{*}{ID} & \multirow{2}{*}{Config.} & \multirow{2}{*}{GAIC} & \multirow{2}{*}{Comp.} & \multirow{2}{*}{Subj.} & GAIC & FLMS & FCDB \\
\cmidrule(lr){6-6}\cmidrule(lr){7-7}\cmidrule(lr){8-8}
 & & & & & SRCC$\uparrow$ & IoU$\uparrow$ & IoU$\uparrow$ \\
\midrule
(a) & None & -- & -- & -- & 0.517 & 0.829 & 0.672 \\
(b) & Direct & -- & -- & -- & 0.526 & 0.812 & 0.671 \\
(c) & Refine & $\checkmark$ & -- & -- & 0.563 & 0.827 & 0.704 \\
(d) & Refine & $\checkmark$ & $\checkmark$ & -- & 0.578 & 0.828 & 0.709 \\
(e) & Refine & $\checkmark$ & $\checkmark$ & $\checkmark$ & \textbf{0.590} & \textbf{0.842} & \textbf{0.714} \\
\bottomrule
\end{tabular*}
\vspace{-0.75\baselineskip}
\vspace{-0.5mm}
\end{wraptable}
\textbf{Justification for off-lattice refinement.} Table~\ref{tab:supp_refinement_scores} compares (a) CPF without refinement, (b) directly using the learned preference score as the reward without refinement, and (c--e) progressively adding aesthetic, composition, and subject-preservation scores for local refinement. The checkmarks indicate scores used for refinement. Relative to the SFT baseline in Table~\ref{tab:ablation}(a), direct optimization of the learned preference score yields modest gains; its GAIC and FLMS improvements are smaller than those of the IoU reward in Table~\ref{tab:ablation}(b), while FCDB IoU is equal. Our refinement uses human annotations to determine preference modes and anchor the local search, while learned preference, regularized by composition and subject-preservation criteria, distinguishes among nearby off-lattice candidates. \textbf{This retains human preference as the foundation and uses visual criteria to improve local targets}, a structure that direct score maximization does not explicitly enforce. Adding composition and subject-preservation regularization in (c--e) improves all three metrics in successive steps. We do not claim to recover the exact human preference landscape; rather, we construct a MOS-grounded approximation and regularize its off-lattice extension to improve alignment with human preference.\par

\medskip
\textbf{Justification for sharpness and structured negatives.} High geometric overlap does not necessarily imply a good composition. An uncropped image or a fixed, oversized template can retain substantial overlap with the ground-truth crop while leaving distracting content or excessive margins. Conversely, a small boundary shift can cut through a subject's head or another visually critical region while barely changing IoU. A preference field that only rewards proximity to positive crops cannot reliably distinguish these failures from acceptable local variations. Our structured negatives explicitly cover such undesirable crops and shape the field to assign lower scores to them, preserving tolerance around good compositions while penalizing visually damaging changes.  Figure~\ref{fig:supp_cpf_scores} demonstrates this distinction: CPF penalizes distracting content, excessive margins, and subject truncation even when IoU remains high. These examples provide compelling qualitative evidence that CPF distinguishes good compositions from perceptually flawed crops in a  manner more consistent with human visual preference than geometric overlap alone.

\begin{figure}[p]
\centering
\input{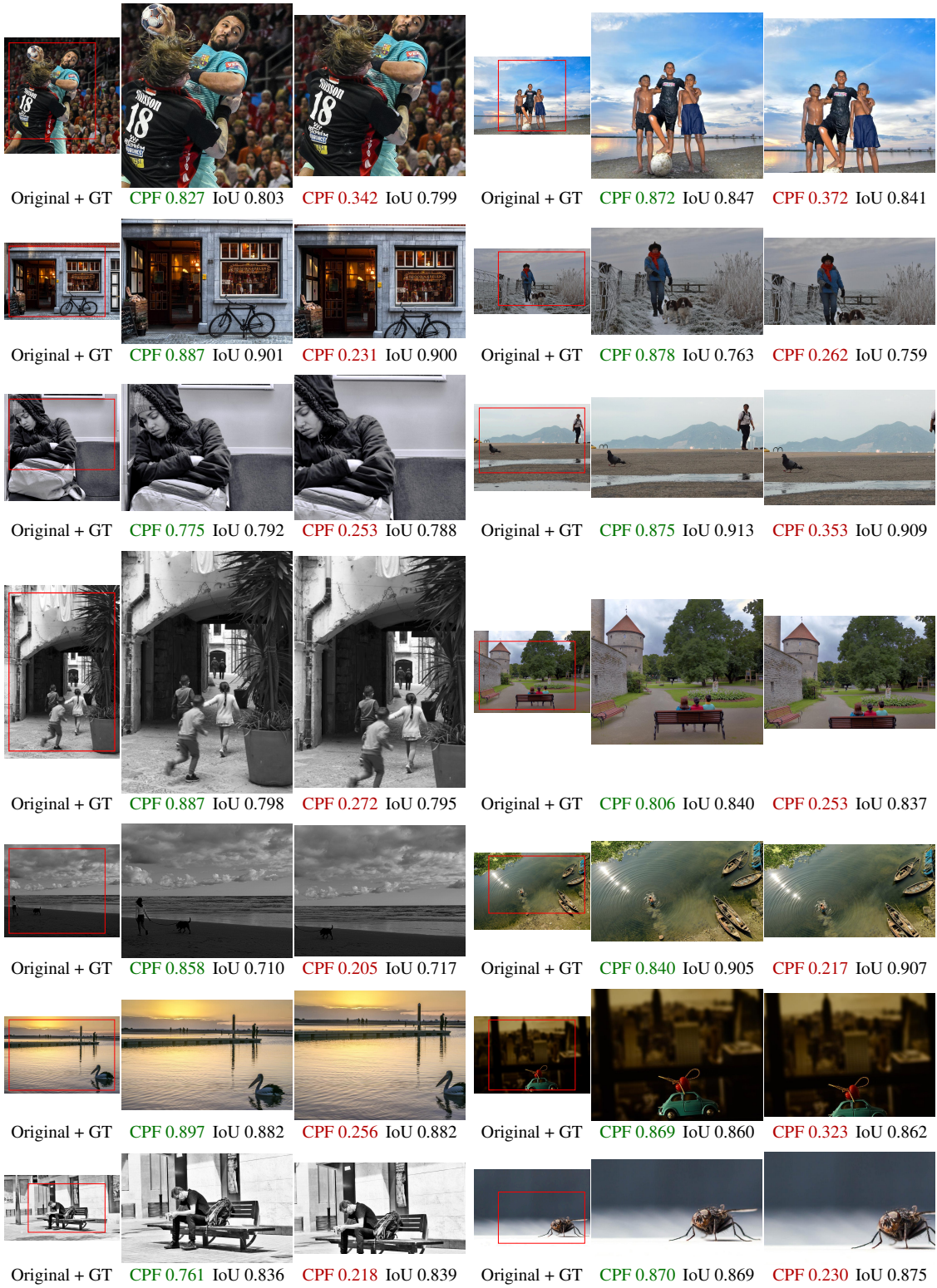}
\caption{CPF scores and IoU for pairs of crops. CPF distinguishes coherent compositions from visually flawed alternatives that IoU fails to adequately penalize, providing qualitative evidence of its closer alignment with human visual preference.}
\label{fig:supp_cpf_scores}
\end{figure}

\textbf{Justification for CPICD.} CPICD is intended to provide better reference annotations for future cropping research, rather than to establish CPIC's superiority through evaluation against its own refinement criterion. Our performance claims are already supported by results on the three established benchmarks, GAIC, FLMS, and FCDB, using their original annotations (Table~\ref{tab:main}). The value of CPICD is instead assessed through human preference: participants favor its recalibrated crops over the original ground truth in Table~\ref{tab:us_recal}, providing evidence beyond agreement with the construction score. By retaining human annotations as anchors while relaxing their discrete sampling constraints and incorporating composition and subject preservation, CPICD offers a more principled construction of reference crops. Our aim is to make annotations better aligned with human satisfaction available as a shared resource for subsequent research.\par

\endgroup

\section{User study details}
\label{sec:supp_user}
\begin{figure}[p]
\centering
\setlength{\abovecaptionskip}{4pt}
\setlength{\belowcaptionskip}{0pt}
\includegraphics[width=0.975\textwidth]{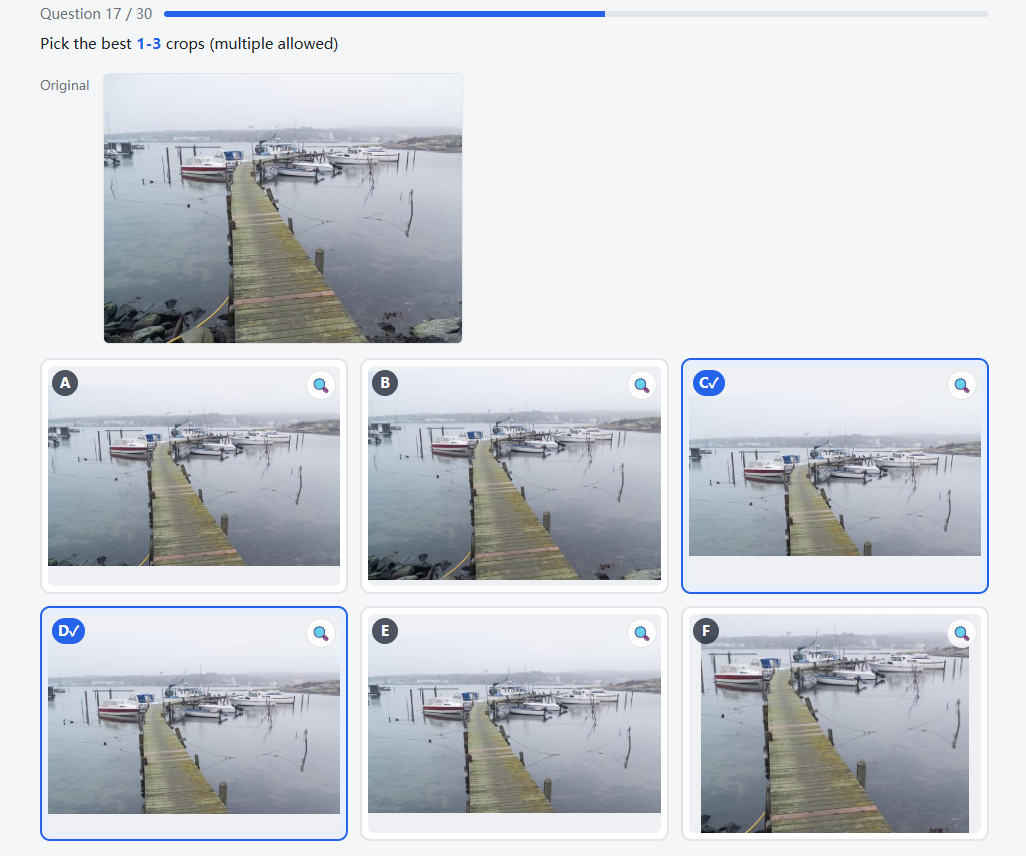}
\caption{User study interface for cropping quality.}
\label{fig:supp_user_ui_crop}
\par\vspace{6pt}
\includegraphics[width=0.975\textwidth]{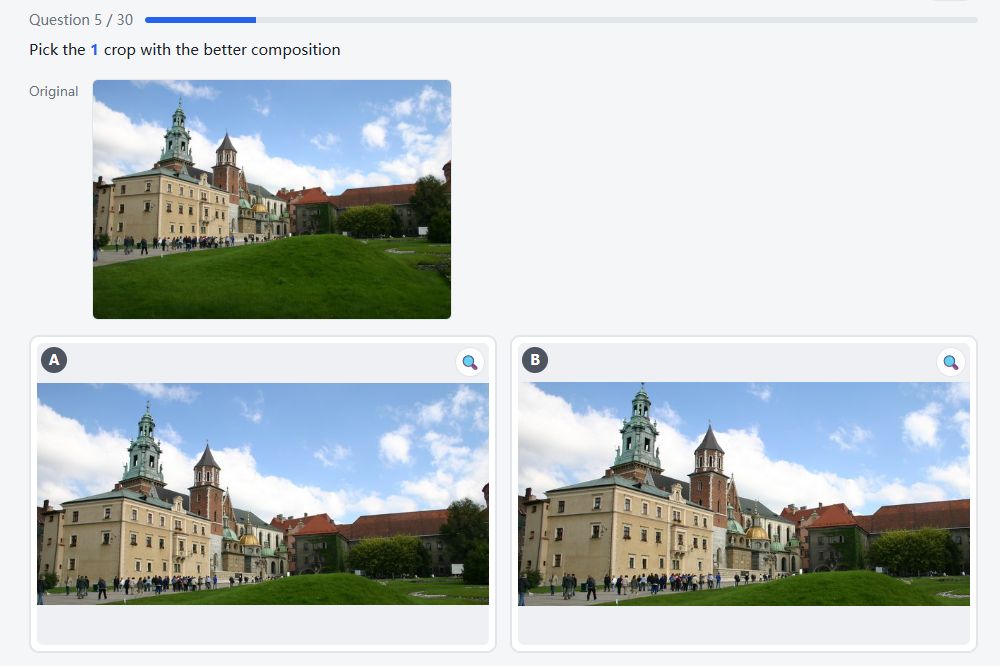}
\caption{User study interface for ground-truth recalibration.}
\label{fig:supp_user_ui_recal}
\end{figure}

We conducted a web-based questionnaire with a total of 1,200 human preference evaluations on cropping quality and ground-truth recalibration (Figures~\ref{fig:supp_user_ui_crop} and~\ref{fig:supp_user_ui_recal}). Twenty participants were instructed to select the most aesthetically pleasing compositions based on their first impressions, with no strictly correct answers. The study size is consistent with prior computer vision user studies~\citep{tang2024lgm,zhang2020deepimageblending,lin2025gdrag,zhang2026infvsr}, while the balanced sampling across datasets and repeated evaluations from multiple participants provide sufficient human judgments for statistically reliable comparisons~\citep{chen2026ntiresr,liu2026ntireinfraredsr,liu2026tacache}. The evaluation comprised two sub-studies, each utilizing 30 test images evenly drawn from GAIC, FLMS, and FCDB (10 per dataset), yielding a total of 1,200 human preference evaluations. For the recalibration study (Table~\ref{tab:us_recal}), users were shown the source image and chose the single best crop between the original benchmark annotation and our CPICD box. For the cropping quality study (Table~\ref{tab:us_crop}), they viewed six anonymized crops (GAIC, CACNet, UNIC, InstructCrop, Venus, and CPIC) and selected their top 1 to 3 choices. To ensure unbiased results, all method names were completely hidden (labeled only A--F), and both the question order and option layouts were randomly shuffled for every participant.

\section{Additional visualizations}
\label{sec:supp_vis}
\begin{figure}[p]
\centering
\includegraphics[width=\textwidth]{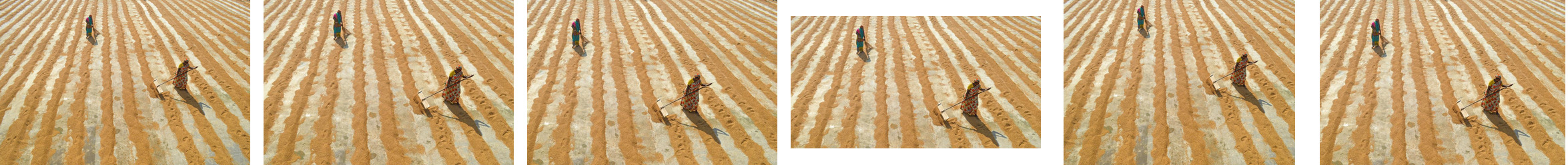}\\[-2.5pt]
\makebox[\textwidth][l]{\scriptsize \makebox[0.159574\textwidth]{Input}\hspace{0.008510\textwidth}\makebox[0.159574\textwidth]{VFN}\hspace{0.008510\textwidth}\makebox[0.159574\textwidth]{VPN}\hspace{0.008510\textwidth}\makebox[0.159574\textwidth]{GAIC}\hspace{0.008510\textwidth}\makebox[0.159574\textwidth]{CACNet}\hspace{0.008510\textwidth}\makebox[0.159574\textwidth]{UNIC}}\\[0pt]
\includegraphics[width=\textwidth]{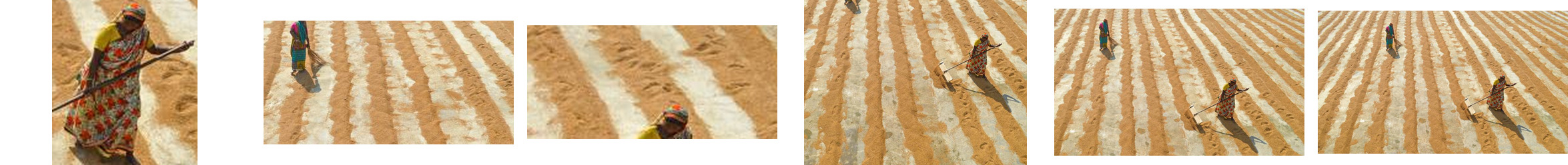}\\[-2.5pt]
\makebox[\textwidth][l]{\scriptsize \makebox[0.159574\textwidth]{Qwen3-VL-4B}\hspace{0.008510\textwidth}\makebox[0.159574\textwidth]{Qwen3-VL-8B}\hspace{0.008510\textwidth}\makebox[0.159574\textwidth]{Qwen2.5-VL-7B}\hspace{0.008510\textwidth}\makebox[0.159574\textwidth]{InstructCrop}\hspace{0.008510\textwidth}\makebox[0.159574\textwidth]{Venus}\hspace{0.008510\textwidth}\makebox[0.159574\textwidth]{\textbf{CPIC (ours)}}}\\[12pt]
\includegraphics[width=\textwidth]{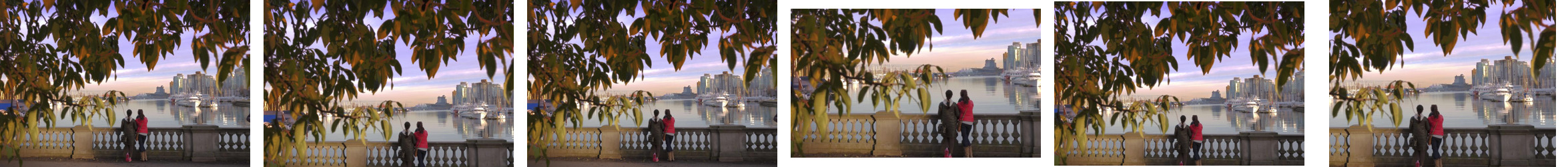}\\[-2.5pt]
\makebox[\textwidth][l]{\scriptsize \makebox[0.159574\textwidth]{Input}\hspace{0.008510\textwidth}\makebox[0.159574\textwidth]{VFN}\hspace{0.008510\textwidth}\makebox[0.159574\textwidth]{VPN}\hspace{0.008510\textwidth}\makebox[0.159574\textwidth]{GAIC}\hspace{0.008510\textwidth}\makebox[0.159574\textwidth]{CACNet}\hspace{0.008510\textwidth}\makebox[0.159574\textwidth]{UNIC}}\\[0pt]
\includegraphics[width=\textwidth]{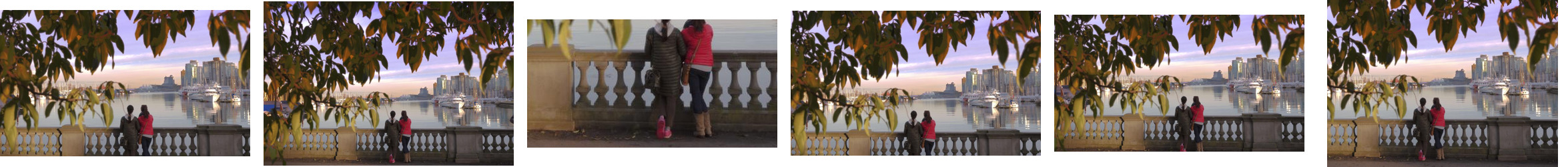}\\[-2.5pt]
\makebox[\textwidth][l]{\scriptsize \makebox[0.159574\textwidth]{Qwen3-VL-8B}\hspace{0.008510\textwidth}\makebox[0.159574\textwidth]{Qwen2.5-VL-7B}\hspace{0.008510\textwidth}\makebox[0.159574\textwidth]{LLaVA-1.5-7B}\hspace{0.008510\textwidth}\makebox[0.159574\textwidth]{InstructCrop}\hspace{0.008510\textwidth}\makebox[0.159574\textwidth]{Venus}\hspace{0.008510\textwidth}\makebox[0.159574\textwidth]{\textbf{CPIC (ours)}}}\\[12pt]
\includegraphics[width=\textwidth]{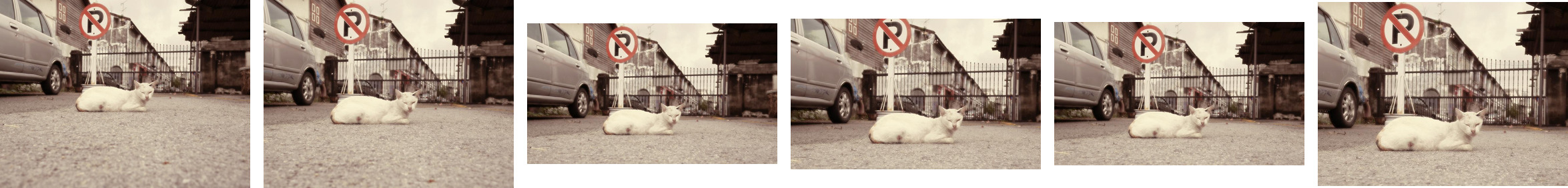}\\[-2.5pt]
\makebox[\textwidth][l]{\scriptsize \makebox[0.159574\textwidth]{Input}\hspace{0.008510\textwidth}\makebox[0.159574\textwidth]{VFN}\hspace{0.008510\textwidth}\makebox[0.159574\textwidth]{VPN}\hspace{0.008510\textwidth}\makebox[0.159574\textwidth]{GAIC}\hspace{0.008510\textwidth}\makebox[0.159574\textwidth]{CACNet}\hspace{0.008510\textwidth}\makebox[0.159574\textwidth]{UNIC}}\\[0pt]
\includegraphics[width=\textwidth]{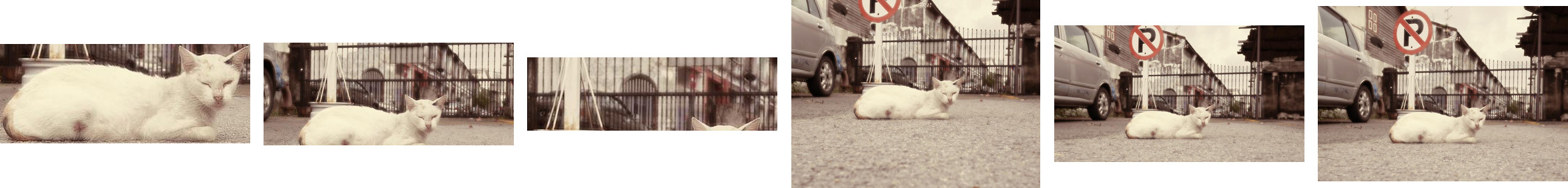}\\[-2.5pt]
\makebox[\textwidth][l]{\scriptsize \makebox[0.159574\textwidth]{Qwen3-VL-4B}\hspace{0.008510\textwidth}\makebox[0.159574\textwidth]{Qwen3-VL-8B}\hspace{0.008510\textwidth}\makebox[0.159574\textwidth]{Qwen2.5-VL-7B}\hspace{0.008510\textwidth}\makebox[0.159574\textwidth]{InstructCrop}\hspace{0.008510\textwidth}\makebox[0.159574\textwidth]{Venus}\hspace{0.008510\textwidth}\makebox[0.159574\textwidth]{\textbf{CPIC (ours)}}}\\[12pt]
\includegraphics[width=\textwidth]{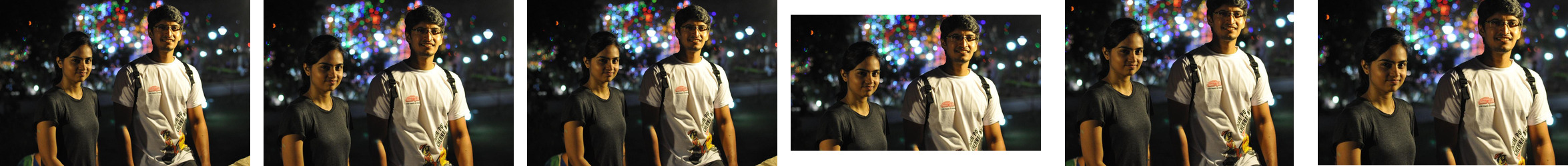}\\[-2.5pt]
\makebox[\textwidth][l]{\scriptsize \makebox[0.159574\textwidth]{Input}\hspace{0.008510\textwidth}\makebox[0.159574\textwidth]{VFN}\hspace{0.008510\textwidth}\makebox[0.159574\textwidth]{VPN}\hspace{0.008510\textwidth}\makebox[0.159574\textwidth]{GAIC}\hspace{0.008510\textwidth}\makebox[0.159574\textwidth]{CACNet}\hspace{0.008510\textwidth}\makebox[0.159574\textwidth]{UNIC}}\\[0pt]
\includegraphics[width=\textwidth]{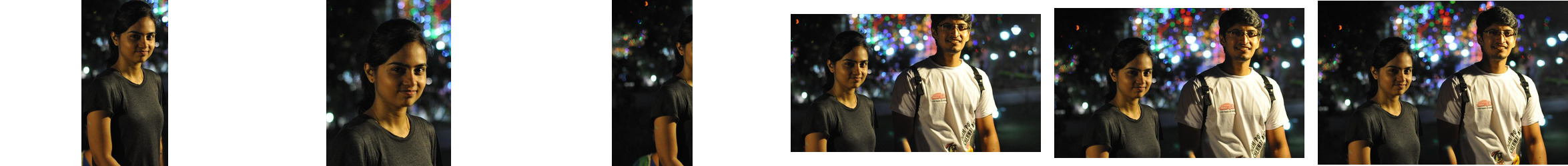}\\[-2.5pt]
\makebox[\textwidth][l]{\scriptsize \makebox[0.159574\textwidth]{Qwen3-VL-4B}\hspace{0.008510\textwidth}\makebox[0.159574\textwidth]{Qwen2.5-VL-7B}\hspace{0.008510\textwidth}\makebox[0.159574\textwidth]{LLaVA-1.5-7B}\hspace{0.008510\textwidth}\makebox[0.159574\textwidth]{InstructCrop}\hspace{0.008510\textwidth}\makebox[0.159574\textwidth]{Venus}\hspace{0.008510\textwidth}\makebox[0.159574\textwidth]{\textbf{CPIC (ours)}}}\\[12pt]
\includegraphics[width=\textwidth]{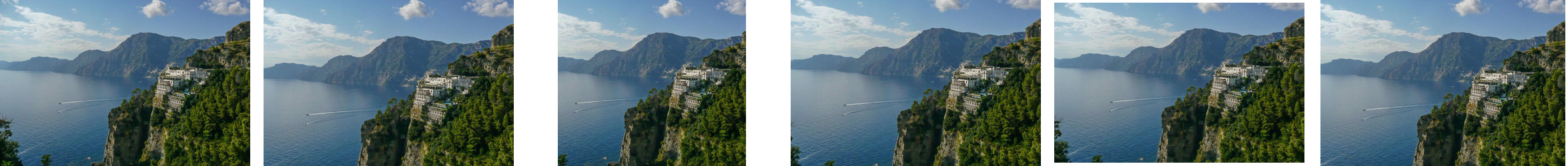}\\[-2.5pt]
\makebox[\textwidth][l]{\scriptsize \makebox[0.159574\textwidth]{Input}\hspace{0.008510\textwidth}\makebox[0.159574\textwidth]{VFN}\hspace{0.008510\textwidth}\makebox[0.159574\textwidth]{VPN}\hspace{0.008510\textwidth}\makebox[0.159574\textwidth]{GAIC}\hspace{0.008510\textwidth}\makebox[0.159574\textwidth]{CACNet}\hspace{0.008510\textwidth}\makebox[0.159574\textwidth]{UNIC}}\\[0pt]
\includegraphics[width=\textwidth]{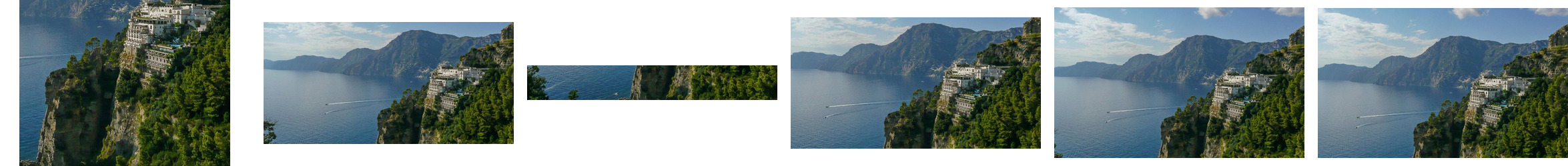}\\[-2.5pt]
\makebox[\textwidth][l]{\scriptsize \makebox[0.159574\textwidth]{Qwen3-VL-4B}\hspace{0.008510\textwidth}\makebox[0.159574\textwidth]{Qwen3-VL-8B}\hspace{0.008510\textwidth}\makebox[0.159574\textwidth]{LLaVA-1.5-7B}\hspace{0.008510\textwidth}\makebox[0.159574\textwidth]{InstructCrop}\hspace{0.008510\textwidth}\makebox[0.159574\textwidth]{Venus}\hspace{0.008510\textwidth}\makebox[0.159574\textwidth]{\textbf{CPIC (ours)}}}
\caption{Additional qualitative comparison (1/4).}
\label{fig:supp_vis_1}
\end{figure}

\begin{figure}[p]
\centering
\includegraphics[width=\textwidth]{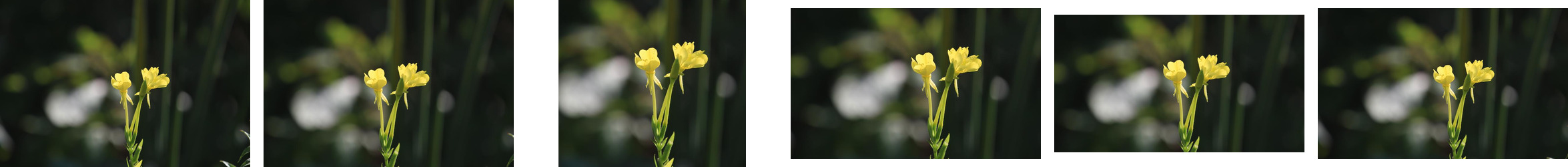}\\[-2.5pt]
\makebox[\textwidth][l]{\scriptsize \makebox[0.159574\textwidth]{Input}\hspace{0.008510\textwidth}\makebox[0.159574\textwidth]{VFN}\hspace{0.008510\textwidth}\makebox[0.159574\textwidth]{VPN}\hspace{0.008510\textwidth}\makebox[0.159574\textwidth]{GAIC}\hspace{0.008510\textwidth}\makebox[0.159574\textwidth]{CACNet}\hspace{0.008510\textwidth}\makebox[0.159574\textwidth]{UNIC}}\\[0pt]
\includegraphics[width=\textwidth]{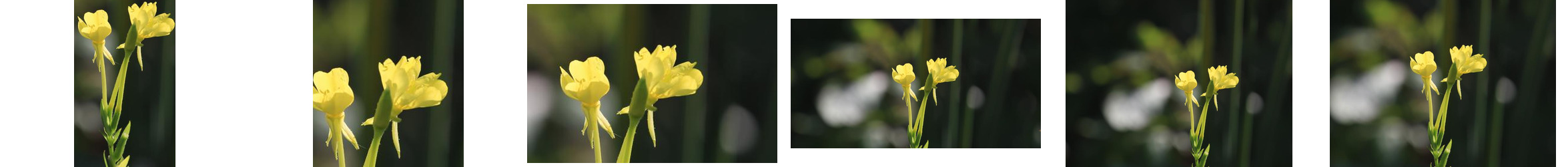}\\[-2.5pt]
\makebox[\textwidth][l]{\scriptsize \makebox[0.159574\textwidth]{Qwen3-VL-4B}\hspace{0.008510\textwidth}\makebox[0.159574\textwidth]{Qwen2.5-VL-7B}\hspace{0.008510\textwidth}\makebox[0.159574\textwidth]{LLaVA-1.5-7B}\hspace{0.008510\textwidth}\makebox[0.159574\textwidth]{InstructCrop}\hspace{0.008510\textwidth}\makebox[0.159574\textwidth]{Venus}\hspace{0.008510\textwidth}\makebox[0.159574\textwidth]{\textbf{CPIC (ours)}}}\\[12pt]
\includegraphics[width=\textwidth]{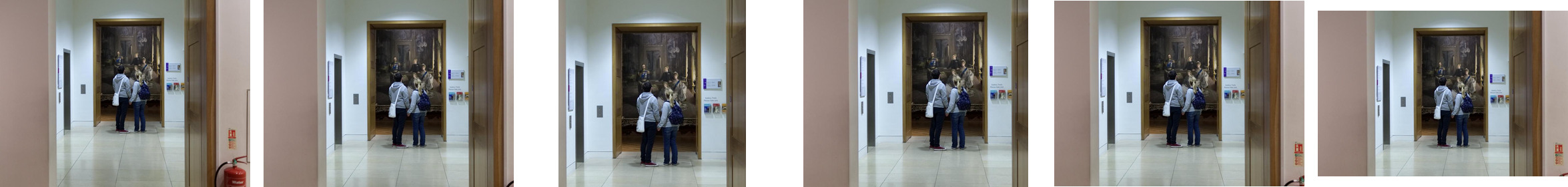}\\[-2.5pt]
\makebox[\textwidth][l]{\scriptsize \makebox[0.159574\textwidth]{Input}\hspace{0.008510\textwidth}\makebox[0.159574\textwidth]{VFN}\hspace{0.008510\textwidth}\makebox[0.159574\textwidth]{VPN}\hspace{0.008510\textwidth}\makebox[0.159574\textwidth]{GAIC}\hspace{0.008510\textwidth}\makebox[0.159574\textwidth]{CACNet}\hspace{0.008510\textwidth}\makebox[0.159574\textwidth]{UNIC}}\\[0pt]
\includegraphics[width=\textwidth]{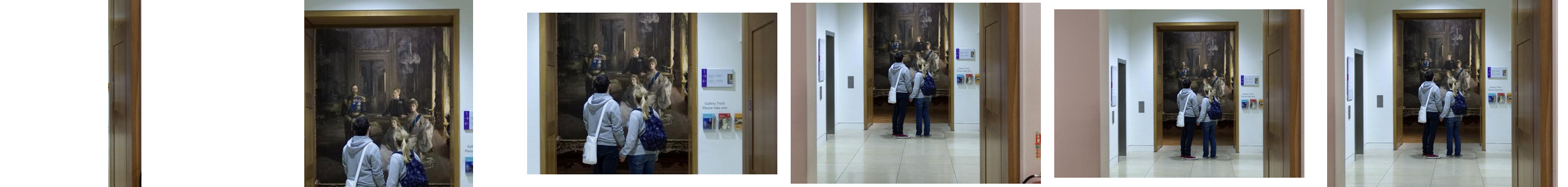}\\[-2.5pt]
\makebox[\textwidth][l]{\scriptsize \makebox[0.159574\textwidth]{Qwen3-VL-2B}\hspace{0.008510\textwidth}\makebox[0.159574\textwidth]{Qwen2.5-VL-7B}\hspace{0.008510\textwidth}\makebox[0.159574\textwidth]{LLaVA-1.5-7B}\hspace{0.008510\textwidth}\makebox[0.159574\textwidth]{InstructCrop}\hspace{0.008510\textwidth}\makebox[0.159574\textwidth]{Venus}\hspace{0.008510\textwidth}\makebox[0.159574\textwidth]{\textbf{CPIC (ours)}}}\\[12pt]
\includegraphics[width=\textwidth]{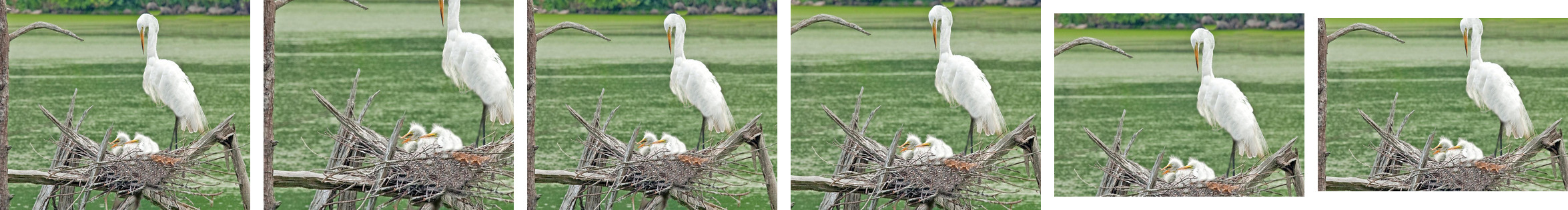}\\[-2.5pt]
\makebox[\textwidth][l]{\scriptsize \makebox[0.159574\textwidth]{Input}\hspace{0.008510\textwidth}\makebox[0.159574\textwidth]{VFN}\hspace{0.008510\textwidth}\makebox[0.159574\textwidth]{VPN}\hspace{0.008510\textwidth}\makebox[0.159574\textwidth]{GAIC}\hspace{0.008510\textwidth}\makebox[0.159574\textwidth]{CACNet}\hspace{0.008510\textwidth}\makebox[0.159574\textwidth]{UNIC}}\\[0pt]
\includegraphics[width=\textwidth]{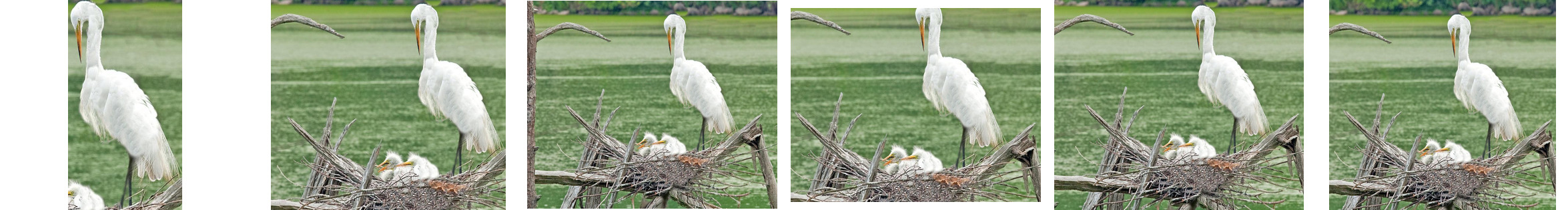}\\[-2.5pt]
\makebox[\textwidth][l]{\scriptsize \makebox[0.159574\textwidth]{Qwen3-VL-2B}\hspace{0.008510\textwidth}\makebox[0.159574\textwidth]{Qwen3-VL-8B}\hspace{0.008510\textwidth}\makebox[0.159574\textwidth]{LLaVA-1.5-7B}\hspace{0.008510\textwidth}\makebox[0.159574\textwidth]{InstructCrop}\hspace{0.008510\textwidth}\makebox[0.159574\textwidth]{Venus}\hspace{0.008510\textwidth}\makebox[0.159574\textwidth]{\textbf{CPIC (ours)}}}\\[12pt]
\includegraphics[width=\textwidth]{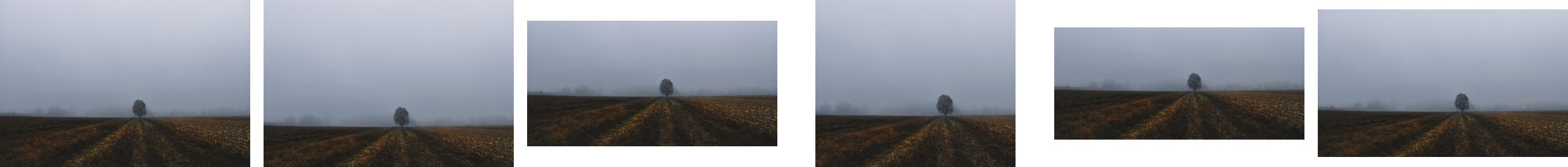}\\[-2.5pt]
\makebox[\textwidth][l]{\scriptsize \makebox[0.159574\textwidth]{Input}\hspace{0.008510\textwidth}\makebox[0.159574\textwidth]{VFN}\hspace{0.008510\textwidth}\makebox[0.159574\textwidth]{VPN}\hspace{0.008510\textwidth}\makebox[0.159574\textwidth]{GAIC}\hspace{0.008510\textwidth}\makebox[0.159574\textwidth]{CACNet}\hspace{0.008510\textwidth}\makebox[0.159574\textwidth]{UNIC}}\\[0pt]
\includegraphics[width=\textwidth]{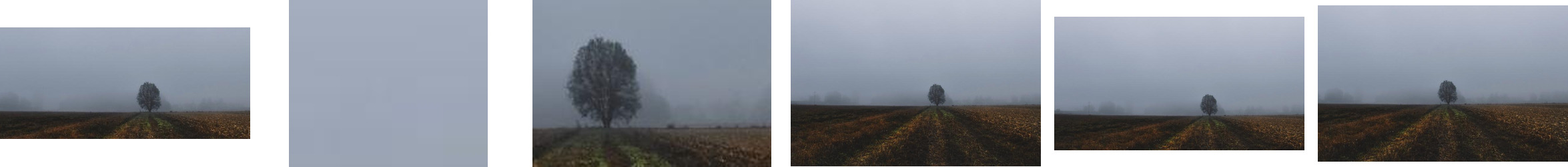}\\[-2.5pt]
\makebox[\textwidth][l]{\scriptsize \makebox[0.159574\textwidth]{Qwen3-VL-8B}\hspace{0.008510\textwidth}\makebox[0.159574\textwidth]{Qwen2.5-VL-7B}\hspace{0.008510\textwidth}\makebox[0.159574\textwidth]{LLaVA-1.5-7B}\hspace{0.008510\textwidth}\makebox[0.159574\textwidth]{InstructCrop}\hspace{0.008510\textwidth}\makebox[0.159574\textwidth]{Venus}\hspace{0.008510\textwidth}\makebox[0.159574\textwidth]{\textbf{CPIC (ours)}}}\\[12pt]
\includegraphics[width=\textwidth]{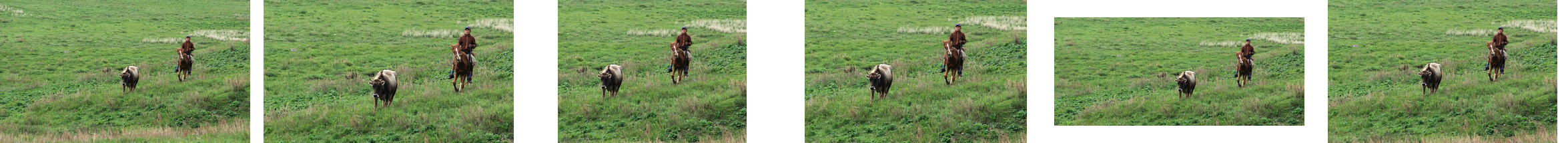}\\[-2.5pt]
\makebox[\textwidth][l]{\scriptsize \makebox[0.159574\textwidth]{Input}\hspace{0.008510\textwidth}\makebox[0.159574\textwidth]{VFN}\hspace{0.008510\textwidth}\makebox[0.159574\textwidth]{VPN}\hspace{0.008510\textwidth}\makebox[0.159574\textwidth]{GAIC}\hspace{0.008510\textwidth}\makebox[0.159574\textwidth]{CACNet}\hspace{0.008510\textwidth}\makebox[0.159574\textwidth]{UNIC}}\\[0pt]
\includegraphics[width=\textwidth]{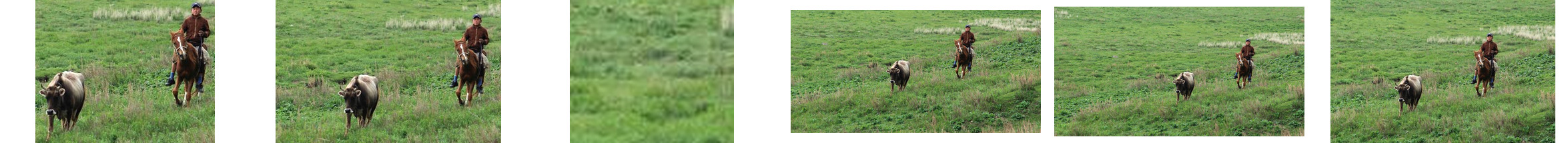}\\[-2.5pt]
\makebox[\textwidth][l]{\scriptsize \makebox[0.159574\textwidth]{Qwen3-VL-4B}\hspace{0.008510\textwidth}\makebox[0.159574\textwidth]{Qwen3-VL-8B}\hspace{0.008510\textwidth}\makebox[0.159574\textwidth]{Qwen2.5-VL-7B}\hspace{0.008510\textwidth}\makebox[0.159574\textwidth]{InstructCrop}\hspace{0.008510\textwidth}\makebox[0.159574\textwidth]{Venus}\hspace{0.008510\textwidth}\makebox[0.159574\textwidth]{\textbf{CPIC (ours)}}}
\caption{Additional qualitative comparison (2/4).}
\label{fig:supp_vis_2}
\end{figure}

\begin{figure}[p]
\centering
\includegraphics[width=\textwidth]{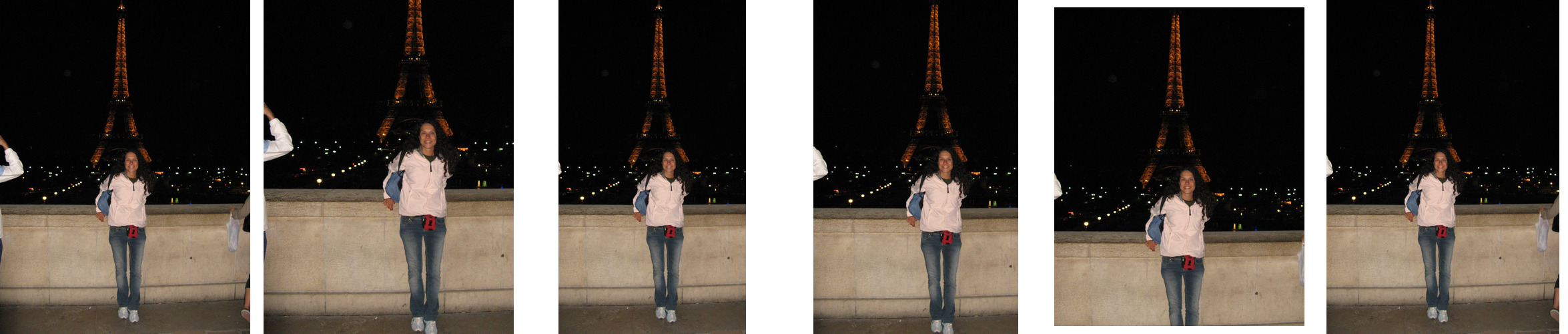}\\[-2.5pt]
\makebox[\textwidth][l]{\scriptsize \makebox[0.159574\textwidth]{Input}\hspace{0.008510\textwidth}\makebox[0.159574\textwidth]{VFN}\hspace{0.008510\textwidth}\makebox[0.159574\textwidth]{VPN}\hspace{0.008510\textwidth}\makebox[0.159574\textwidth]{GAIC}\hspace{0.008510\textwidth}\makebox[0.159574\textwidth]{CACNet}\hspace{0.008510\textwidth}\makebox[0.159574\textwidth]{UNIC}}\\[0pt]
\includegraphics[width=\textwidth]{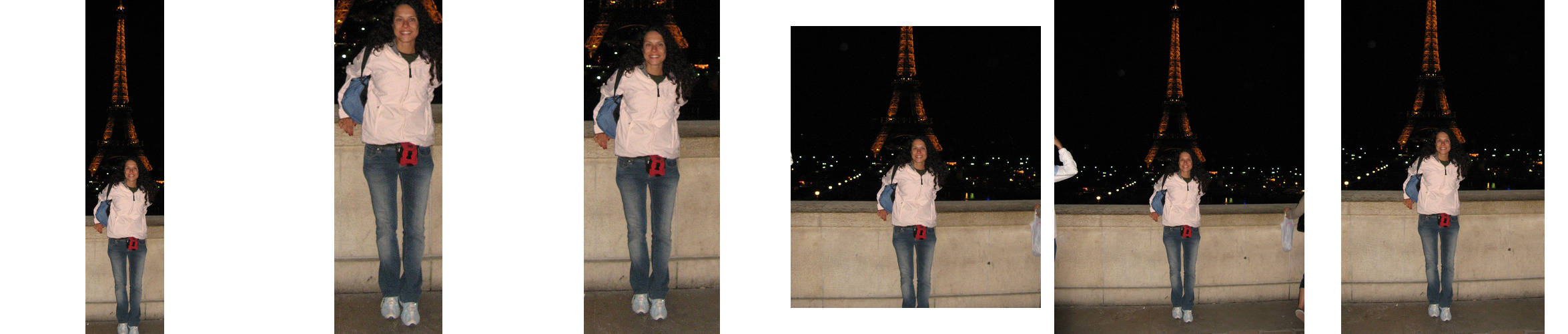}\\[-2.5pt]
\makebox[\textwidth][l]{\scriptsize \makebox[0.159574\textwidth]{Qwen3-VL-8B}\hspace{0.008510\textwidth}\makebox[0.159574\textwidth]{Qwen2.5-VL-7B}\hspace{0.008510\textwidth}\makebox[0.159574\textwidth]{LLaVA-1.5-7B}\hspace{0.008510\textwidth}\makebox[0.159574\textwidth]{InstructCrop}\hspace{0.008510\textwidth}\makebox[0.159574\textwidth]{Venus}\hspace{0.008510\textwidth}\makebox[0.159574\textwidth]{\textbf{CPIC (ours)}}}\\[12pt]
\includegraphics[width=\textwidth]{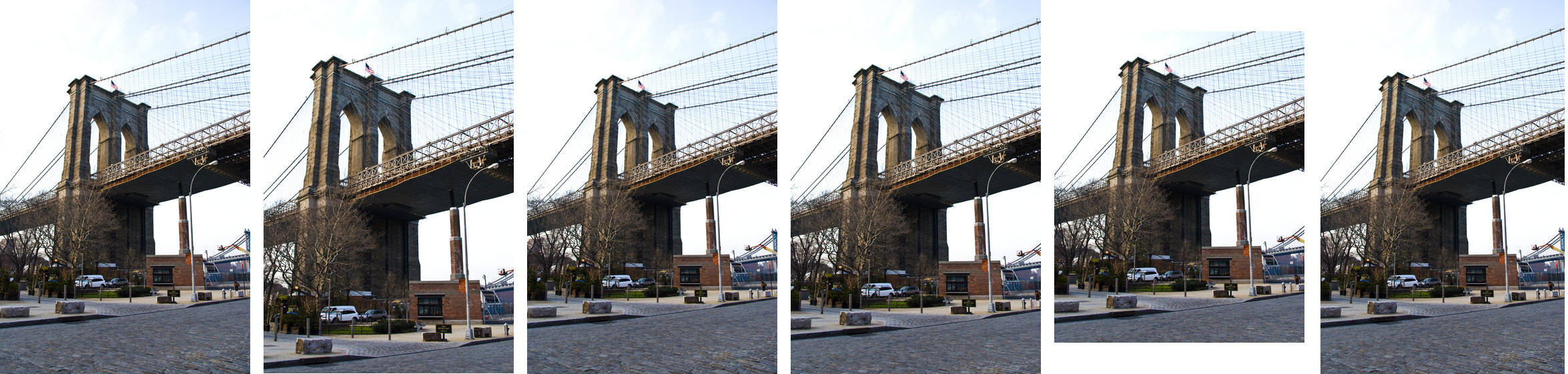}\\[-2.5pt]
\makebox[\textwidth][l]{\scriptsize \makebox[0.159574\textwidth]{Input}\hspace{0.008510\textwidth}\makebox[0.159574\textwidth]{VFN}\hspace{0.008510\textwidth}\makebox[0.159574\textwidth]{VPN}\hspace{0.008510\textwidth}\makebox[0.159574\textwidth]{GAIC}\hspace{0.008510\textwidth}\makebox[0.159574\textwidth]{CACNet}\hspace{0.008510\textwidth}\makebox[0.159574\textwidth]{UNIC}}\\[0pt]
\includegraphics[width=\textwidth]{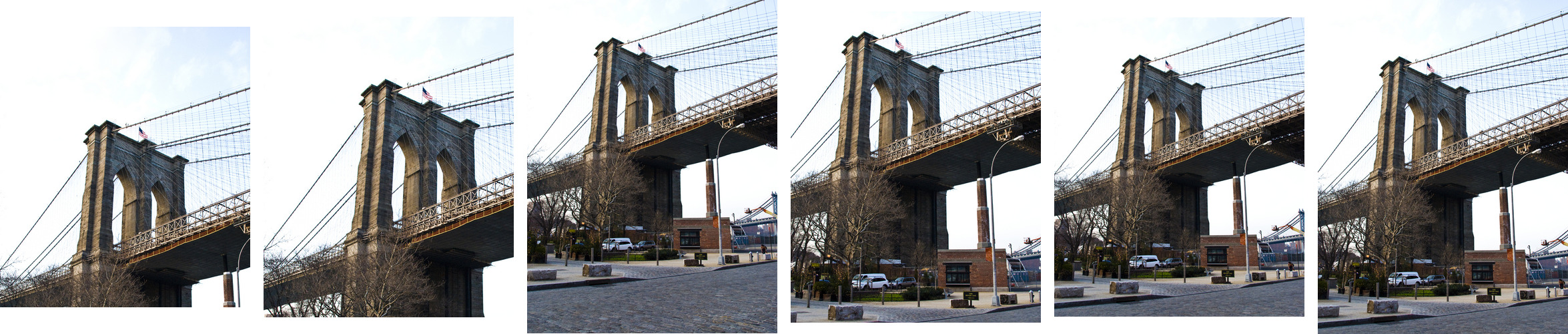}\\[-2.5pt]
\makebox[\textwidth][l]{\scriptsize \makebox[0.159574\textwidth]{Qwen3-VL-8B}\hspace{0.008510\textwidth}\makebox[0.159574\textwidth]{Qwen2.5-VL-7B}\hspace{0.008510\textwidth}\makebox[0.159574\textwidth]{LLaVA-1.5-7B}\hspace{0.008510\textwidth}\makebox[0.159574\textwidth]{InstructCrop}\hspace{0.008510\textwidth}\makebox[0.159574\textwidth]{Venus}\hspace{0.008510\textwidth}\makebox[0.159574\textwidth]{\textbf{CPIC (ours)}}}\\[12pt]
\includegraphics[width=\textwidth]{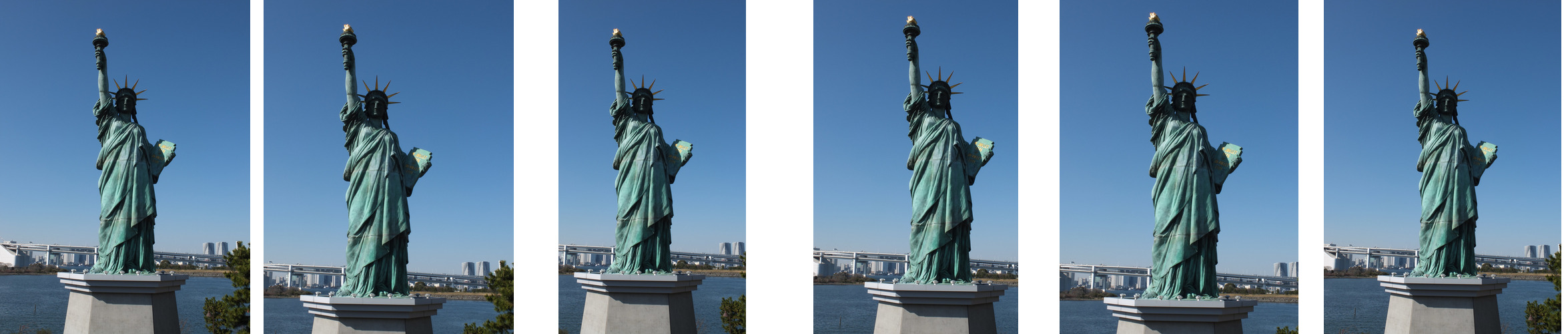}\\[-2.5pt]
\makebox[\textwidth][l]{\scriptsize \makebox[0.159574\textwidth]{Input}\hspace{0.008510\textwidth}\makebox[0.159574\textwidth]{VFN}\hspace{0.008510\textwidth}\makebox[0.159574\textwidth]{VPN}\hspace{0.008510\textwidth}\makebox[0.159574\textwidth]{GAIC}\hspace{0.008510\textwidth}\makebox[0.159574\textwidth]{CACNet}\hspace{0.008510\textwidth}\makebox[0.159574\textwidth]{UNIC}}\\[0pt]
\includegraphics[width=\textwidth]{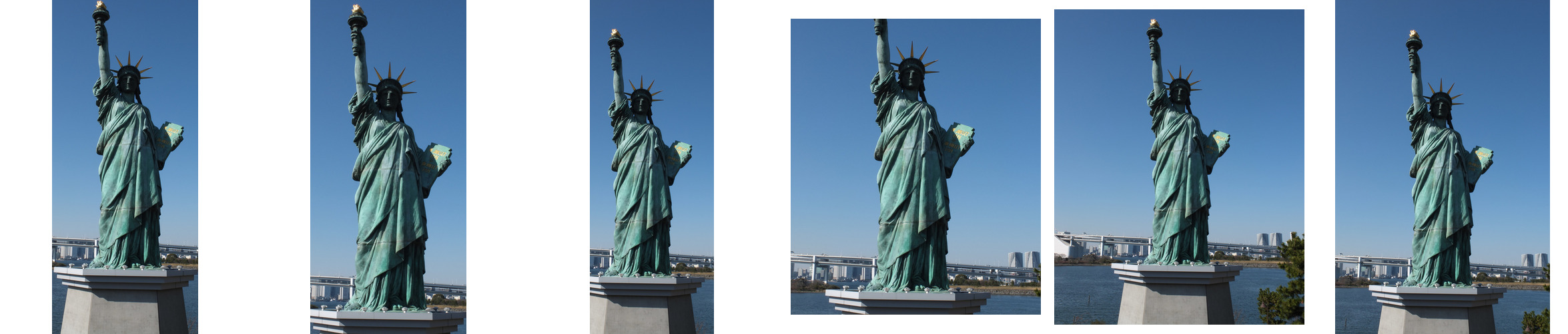}\\[-2.5pt]
\makebox[\textwidth][l]{\scriptsize \makebox[0.159574\textwidth]{Qwen3-VL-4B}\hspace{0.008510\textwidth}\makebox[0.159574\textwidth]{Qwen3-VL-8B}\hspace{0.008510\textwidth}\makebox[0.159574\textwidth]{LLaVA-1.5-7B}\hspace{0.008510\textwidth}\makebox[0.159574\textwidth]{InstructCrop}\hspace{0.008510\textwidth}\makebox[0.159574\textwidth]{Venus}\hspace{0.008510\textwidth}\makebox[0.159574\textwidth]{\textbf{CPIC (ours)}}}
\caption{Additional qualitative comparison (3/4).}
\label{fig:supp_vis_3}
\end{figure}

\begin{figure}[p]
\centering
\includegraphics[width=\textwidth]{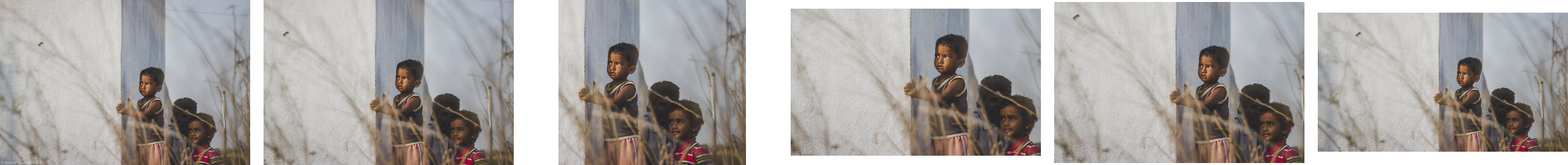}\\[-2.5pt]
\makebox[\textwidth][l]{\scriptsize \makebox[0.159574\textwidth]{Input}\hspace{0.008510\textwidth}\makebox[0.159574\textwidth]{VFN}\hspace{0.008510\textwidth}\makebox[0.159574\textwidth]{VPN}\hspace{0.008510\textwidth}\makebox[0.159574\textwidth]{GAIC}\hspace{0.008510\textwidth}\makebox[0.159574\textwidth]{CACNet}\hspace{0.008510\textwidth}\makebox[0.159574\textwidth]{UNIC}}\\[0pt]
\includegraphics[width=\textwidth]{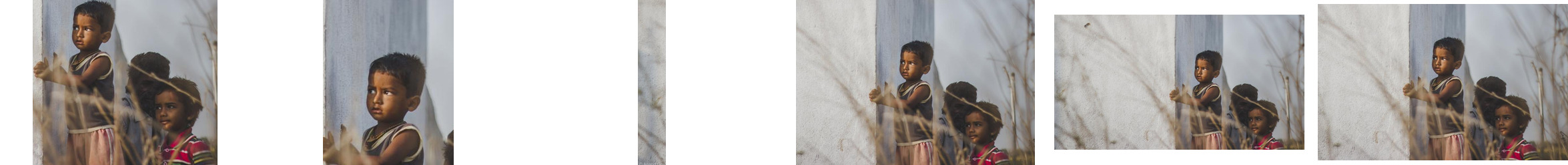}\\[-2.5pt]
\makebox[\textwidth][l]{\scriptsize \makebox[0.159574\textwidth]{Qwen3-VL-2B}\hspace{0.008510\textwidth}\makebox[0.159574\textwidth]{Qwen2.5-VL-7B}\hspace{0.008510\textwidth}\makebox[0.159574\textwidth]{LLaVA-1.5-7B}\hspace{0.008510\textwidth}\makebox[0.159574\textwidth]{InstructCrop}\hspace{0.008510\textwidth}\makebox[0.159574\textwidth]{Venus}\hspace{0.008510\textwidth}\makebox[0.159574\textwidth]{\textbf{CPIC (ours)}}}\\[12pt]
\includegraphics[width=\textwidth]{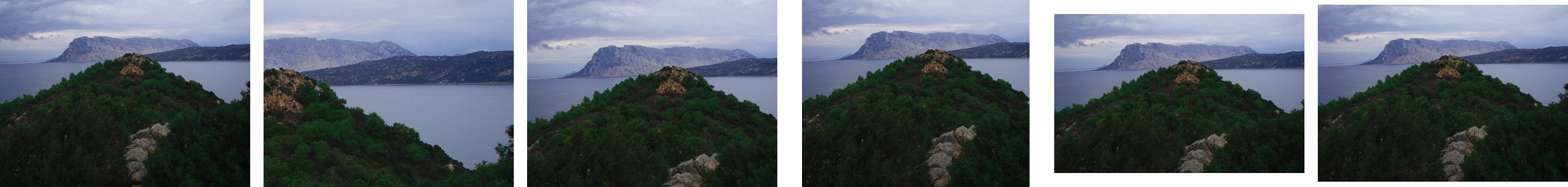}\\[-2.5pt]
\makebox[\textwidth][l]{\scriptsize \makebox[0.159574\textwidth]{Input}\hspace{0.008510\textwidth}\makebox[0.159574\textwidth]{VFN}\hspace{0.008510\textwidth}\makebox[0.159574\textwidth]{VPN}\hspace{0.008510\textwidth}\makebox[0.159574\textwidth]{GAIC}\hspace{0.008510\textwidth}\makebox[0.159574\textwidth]{CACNet}\hspace{0.008510\textwidth}\makebox[0.159574\textwidth]{UNIC}}\\[0pt]
\includegraphics[width=\textwidth]{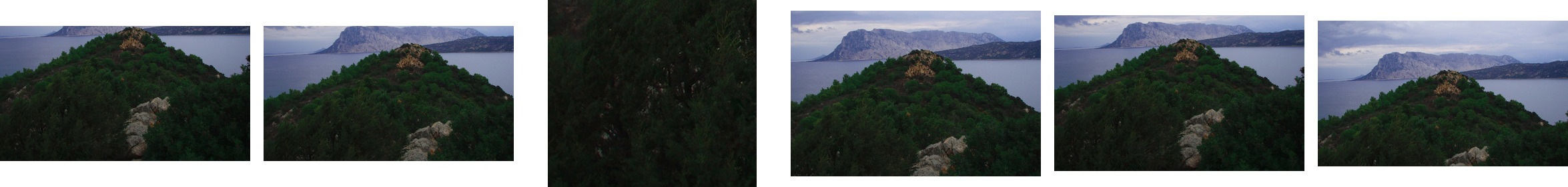}\\[-2.5pt]
\makebox[\textwidth][l]{\scriptsize \makebox[0.159574\textwidth]{Qwen3-VL-4B}\hspace{0.008510\textwidth}\makebox[0.159574\textwidth]{Qwen3-VL-8B}\hspace{0.008510\textwidth}\makebox[0.159574\textwidth]{LLaVA-1.5-7B}\hspace{0.008510\textwidth}\makebox[0.159574\textwidth]{InstructCrop}\hspace{0.008510\textwidth}\makebox[0.159574\textwidth]{Venus}\hspace{0.008510\textwidth}\makebox[0.159574\textwidth]{\textbf{CPIC (ours)}}}\\[12pt]
\includegraphics[width=\textwidth]{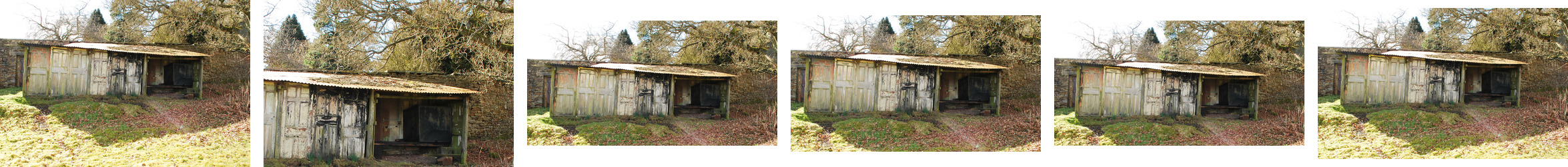}\\[-2.5pt]
\makebox[\textwidth][l]{\scriptsize \makebox[0.159574\textwidth]{Input}\hspace{0.008510\textwidth}\makebox[0.159574\textwidth]{VFN}\hspace{0.008510\textwidth}\makebox[0.159574\textwidth]{VPN}\hspace{0.008510\textwidth}\makebox[0.159574\textwidth]{GAIC}\hspace{0.008510\textwidth}\makebox[0.159574\textwidth]{CACNet}\hspace{0.008510\textwidth}\makebox[0.159574\textwidth]{UNIC}}\\[0pt]
\includegraphics[width=\textwidth]{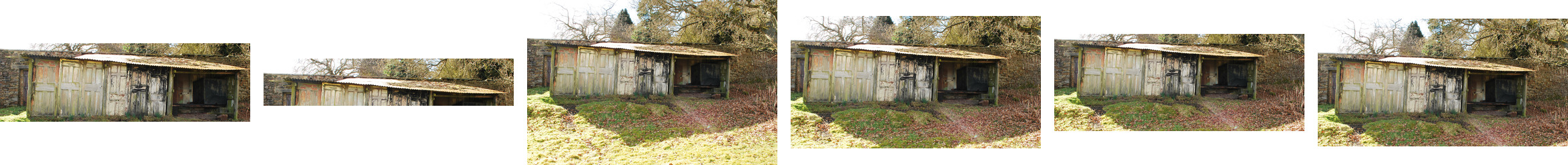}\\[-2.5pt]
\makebox[\textwidth][l]{\scriptsize \makebox[0.159574\textwidth]{Qwen3-VL-8B}\hspace{0.008510\textwidth}\makebox[0.159574\textwidth]{Qwen2.5-VL-7B}\hspace{0.008510\textwidth}\makebox[0.159574\textwidth]{LLaVA-1.5-7B}\hspace{0.008510\textwidth}\makebox[0.159574\textwidth]{InstructCrop}\hspace{0.008510\textwidth}\makebox[0.159574\textwidth]{Venus}\hspace{0.008510\textwidth}\makebox[0.159574\textwidth]{\textbf{CPIC (ours)}}}
\caption{Additional qualitative comparison (4/4).}
\label{fig:supp_vis_4}
\end{figure}

\begin{figure}[p]
\centering
\input{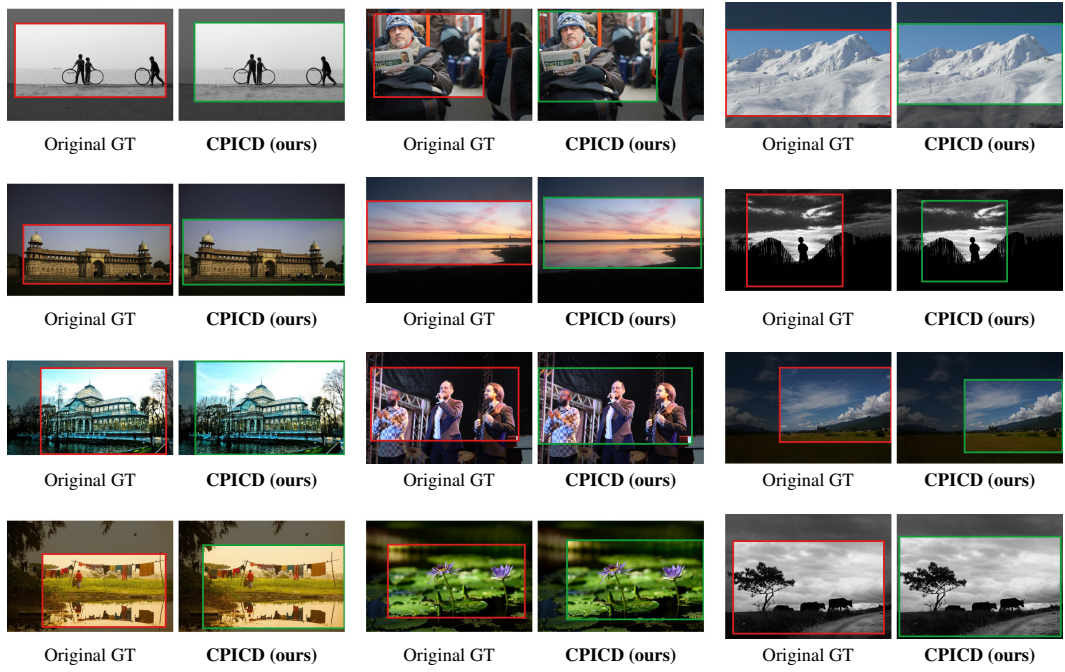}
\caption{Original ground truth (red) vs.\ our recalibrated boxes (green).}
\label{fig:supp_recal}
\end{figure}

To further demonstrate the aesthetic superiority of our method, Figures~\ref{fig:supp_vis_1}--\ref{fig:supp_vis_4} provide additional qualitative comparisons against traditional networks, zero-shot VLMs, and VLM-based croppers using test images from GAIC, FCDB and FLMS. Furthermore, to illustrate the superior quality of our recalibrated benchmark, Figure~\ref{fig:supp_recal} visualizes  supplementary examples (Table~\ref{tab:us_recal}). These visual pairs highlight instances across all three datasets where our CPICD recalibrated boxes (green) visibly excel the original annotations (red).

\end{document}